\documentclass[10pt,twocolumn,letterpaper]{article}

\usepackage[pagenumbers]{wacv} % To force page numbers, e.g. for an arXiv version

\usepackage{graphicx}
\usepackage{amsmath}
\usepackage{amssymb}
\usepackage{booktabs}
\usepackage[export]{adjustbox}
\usepackage{epsfig}
\usepackage{algorithm}
\usepackage{algpseudocode}
\usepackage{enumitem}
\usepackage{color, colortbl}
\usepackage{multirow}
\usepackage{url}
\usepackage{enumitem}
\usepackage[accsupp]{axessibility}  % Improves PDF readability for those with disabilities.

\usepackage[pagebackref,breaklinks,colorlinks]{hyperref}

\usepackage{cuted}
\usepackage{capt-of}

\usepackage[capitalize]{cleveref}
\crefname{section}{Sec.}{Secs.}
\Crefname{section}{Section}{Sections}
\Crefname{table}{Table}{Tables}
\crefname{table}{Tab.}{Tabs.}
\definecolor{myblue}{RGB}{0,0,128}
\definecolor{mygray}{RGB}{220,220,220}

\newcommand\blfootnote[1]{%
  \begingroup
  \renewcommand\thefootnote{}\footnote{#1}%
  \addtocounter{footnote}{-1}%
  \endgroup
}

\def\wacvPaperID{2365} % *** Enter the WACV Paper ID here
\def\confName{WACV}
\def\confYear{2024}
\newcommand*{\img}[1]{%
    \raisebox{-.15\baselineskip}{%
        \includegraphics[
        height=\baselineskip,
        width=\baselineskip,
        keepaspectratio,
        ]{#1}%
    }%
}

\newcommand{\vdgr}{$\mathbb{VD}$-$\mathbb{GR}$}

\title{
\vdgr: Boosting ${\mathbb{V}}$isual ${\mathbb{D}}$ialog with Cascaded Spatial-Temporal\\Multi-Modal ${\mathbb{GR}}$aphs}

\author{Adnen Abdessaied \quad  \quad Lei Shi \quad \quad  Andreas Bulling\\
University of Stuttgart\\
{\tt\small \{adnen.abdessaied, lei.shi, andreas.bulling\}@vis.uni-stuttgart.de}
}

\begin{document}
\maketitle

\begin{strip}\centering
\includegraphics[width=0.94\textwidth]{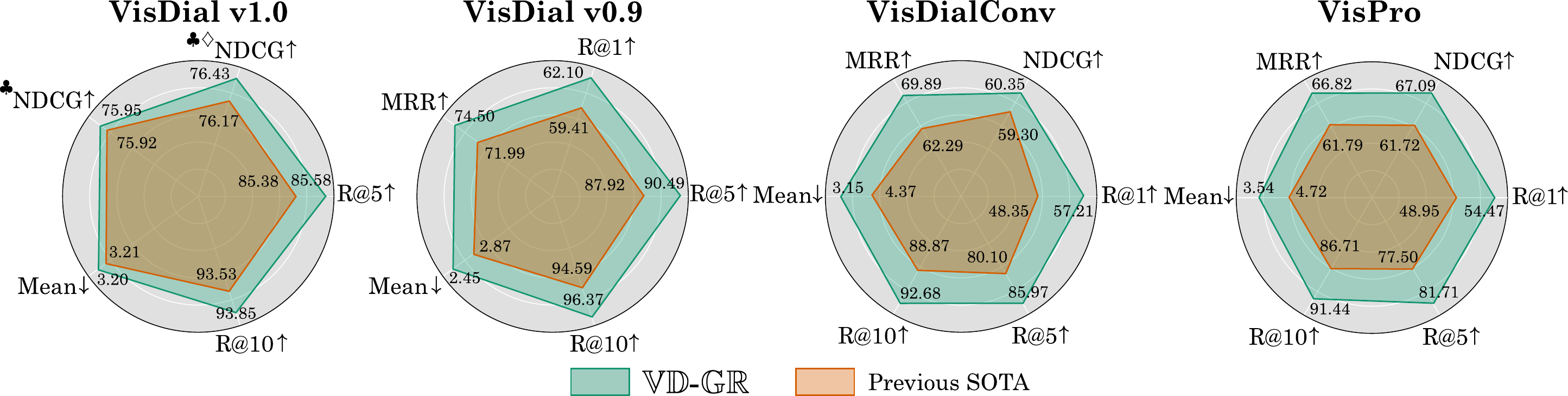}
\captionof{figure}{\vdgr\, outperforms strong baselines and achieves new state-of-the-art results on VisDial v1.0, VisDial v0.9, VisDialConv, and VisPro.
$\uparrow$ indicates higher is better and $\downarrow$ indicates lower is better.
($\clubsuit = $ Fine-tuning on dense annotations, 
$\diamondsuit = $ Ensemble model). 
}
\label{fig:teaser}
\end{strip}

\begin{abstract}
We propose \vdgr \blfootnote{\img{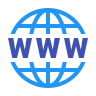} Our project web-page is accessible \href{https://perceptualui.org/publications/abdessaied24_wacv/}{here}.} -- a novel visual dialog model that combines pre-trained language models (LMs) with graph neural networks (GNNs).
Prior works mainly focused on one class of models at the expense of the other, thus missing out on the opportunity of combining their respective benefits.
At the core of \vdgr\, is a novel integration mechanism that alternates between spatial-temporal multi-modal GNNs and BERT layers, and that covers three distinct contributions:
First, we use multi-modal GNNs to process the features of each modality (image, question, and dialog history) and exploit their local structures before performing BERT global attention.
Second, we propose hub-nodes that link to all other nodes within one modality graph, allowing the model to propagate information from one GNN (modality) to the other in a cascaded manner.
Third, we augment the BERT hidden states with fine-grained multi-modal GNN features before passing them to the next \vdgr\, layer.
Evaluations on VisDial v1.0, VisDial v0.9, VisDialConv, and VisPro show that \vdgr\, achieves new state-of-the-art results across all four datasets.
\end{abstract}
\section{Introduction}
\label{sec:intro}

Visual dialog is a multi-modal task to assess how well an artificial agent can hold a conversation with a human on a visual content using natural language \cite{visdial}.
Visual dialog differs from other tasks, such as visual \cite{VQA} or video question answering \cite{video_qa}, in that it requires the agent to answer a series of \textit{temporally dependent} questions.
That is, the agent not only has to reason about the visual input but also has to leverage the context of previous rounds to be able to answer the current question correctly.
Although other datasets have been proposed for this task \cite{NIPS2017_654ad60e, Kottur2019}, VisDial \cite{visdial} has established itself as the de-facto standard because of its challenging, open-ended, and real-world nature.

Early visual dialog models on this dataset were based on recurrent networks \cite{rnn} within deep neural architectures ranging from vanilla LSTMs \cite{hochreiter1997long} over memory nets \cite{mem_nets} to hierarchical structures \cite{hierarchical}.
More recently, graph neural networks (GNNs) %\cite{gnns}
have been proposed and have been shown to produce more fine-grained features based on the local structures of each modality \cite{gog, Guo2020, jiang2020visual, li2019relation, kbgn}.
Other works have focused on attention models and pre-trained language models (LM) \cite{visdial,Schwartz2019,Nguyen2020, Agarwal2020}, fine-tuned for the visual dialog task
\cite{murahari2020large, li-moens-2021-modeling, Wang2020, vd_pcr}.
However, both methodological approaches have so far remained separate despite their
complementary strengths and weaknesses:
While GNNs are effective at exploiting local structure,
they struggle to capture the global inter-modal context, especially in a rich multi-modal task like visual dialog.
In contrast, transformer-based models \cite{transformer} (e.g. BERT \cite{bert})
excel at learning the global context using self and global attention
but often fail to exploit local intra-modal structures within each modality and suffer from the lack of inductive bias \cite{xu2021vitae}. 

We posit that
it is essential to exploit both local intra-modal structures and the global inter-modal context:
Each modality is composed of smaller entities whose relationships have to be separately captured and understood by the model (i.e. the objects in the image, the words that constitute the question, and the rounds that form the history).
At the same time, the inter-dependency of the modalities and the global context are equally important for a model to be able to answer the current question efficiently.
Implementing this idea, we propose \vdgr\, -- the first visual dialog model to combine transformer-based LMs and GNNs.
Each of our proposed layers alternates between GNNs that use hub-nodes to propagate information from one modality graph to another to alleviate the lack of inter-modal context, as well as BERT layers to learn the global intra-model context.
The contributions of our work are threefold:
\begin{itemize}[wide,labelindent=0.0cm,leftmargin=0.0cm]
 \item A novel integration method of GNNs into transformer-based models that alternates between multi-modal graph aggregation and BERT layers.
The GNNs exploit the local structure of each modality to augment the BERT hidden states with their fine-grained features before passing them to the next \vdgr\, layer in an attempt to mitigate their lack of inductive bias.
\item A novel feature propagation technique for multi-modal GNNs that relies on hub-nodes that link to all other nodes of the other modality in a cascaded manner, thus alleviating the lack of inter-modal context within the multi-modal graphs.
\item Our model achieves new state-of-the-art results on VisDial v1.0, VisDial v0.9, VisDialConv, and VisPro datasets, thereby outperforming strong baselines on all four datasets.

\end{itemize}

\begin{figure*}[ht]
    \begin{minipage}{1\linewidth}
        \centering
        \scalebox{0.99}[0.99]{
            \includegraphics[width=\textwidth]{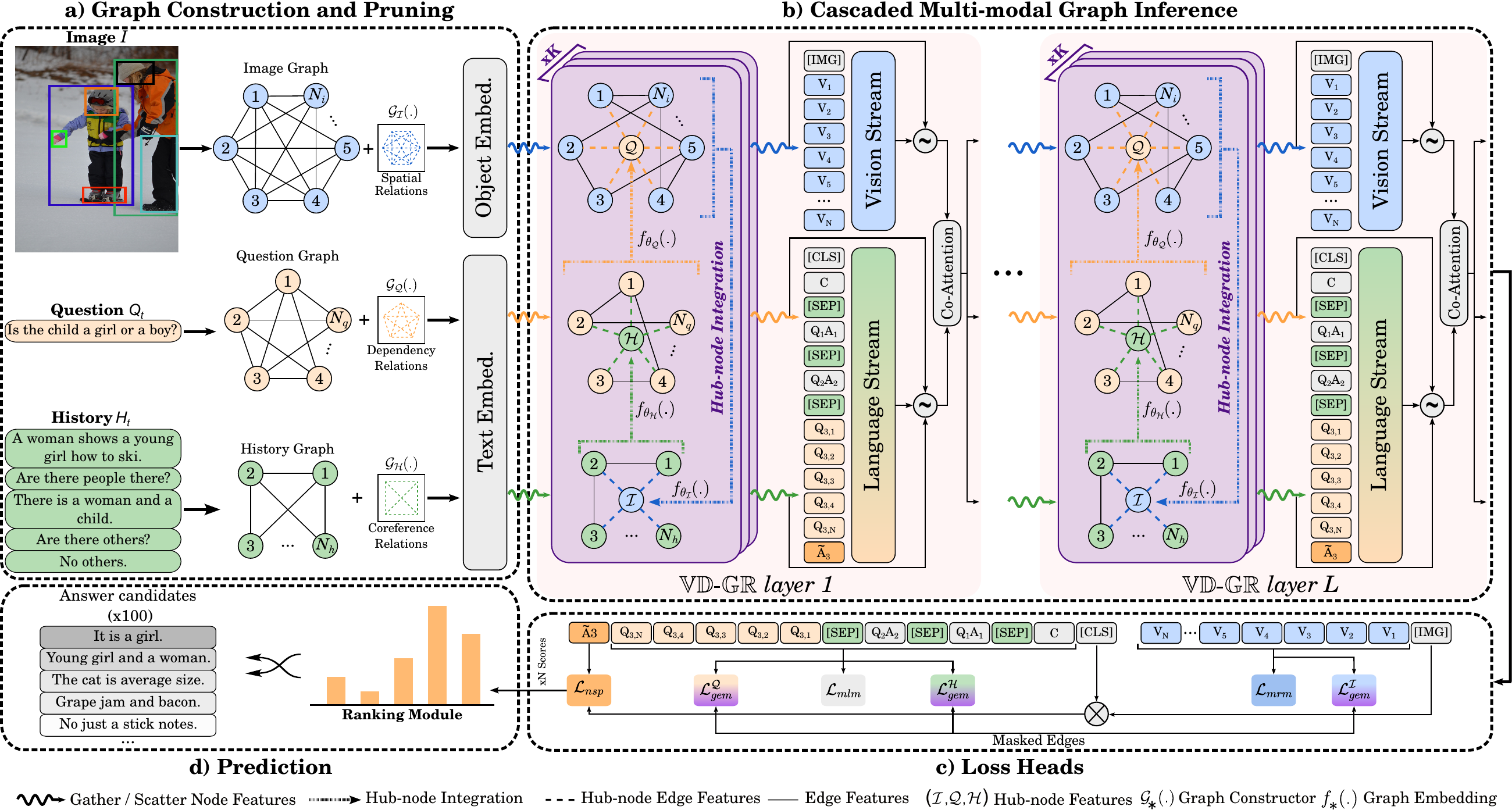}
        }
        \caption{
            \vdgr\, takes an image, a question, and a history consisting of the previous turns as input together with their respective graph structures.
            Each \vdgr-layer is composed of cascaded multi-modal graphs and a vanilla BERT layer.
        }
        \label{fig:method}
    \end{minipage}
\end{figure*}

\section{Related Work}
\label{sec:related_work}
\subsection{GNNs for Visual Dialog}
GNN-EM \cite{zheng2019reasoning} was one of the earliest models to deploy GNNs for visual dialog.
It relied on building graphically structured neural networks to approximate the learning and inference processes of graphical models \cite{arxiv.1506.02108,sukhbaatar2016learning,arxiv.1511.05493,arxiv.1710.06513}.
FGA \cite{Schwartz2019} developed a general attention mechanism that borrowed from the core idea of GNN message passing, and that was able to combine representations of any number of utilities.
CAG \cite{Guo2020} proposed representing the image as a fully-connected graph of objects whose adjacency matrices could be dynamically updated using a question-based attention mechanism.
GOG \cite{gog} proposed representing each modality as a graph before processing them by a light-weight fusion module \cite{Nguyen2020} to rank the candidate answers.
Similarly to CAG \cite{Guo2020}, DualVD \cite{jiang2019dualvd} proposed representing the image as a graph consisting of connected objects. They argued that the visual view helped to capture the appearance-level information, including objects and their relationships, while the semantic view enabled the agent to understand high-level visual semantics from the whole image to the local regions \cite{jiang2019dualvd}.

Our model differs from the aforementioned works in two major aspects:
(1) Instead of concatenating the GNN node features, we propose using hub-nodes that link to all remaining nodes within a given modality.
We argue that this improves the feature fusion on a local scale given that GNNs can freely learn how to integrate information extracted from previous modalities.
(2) We marry GNNs with pre-trained LMs (BERT) in a novel way to benefit from the advantages of both worlds.
Contrarily to \cite{Yang2021,ying2021do} where simple integration approaches for plain uni-modal tasks were introduced, we propose a sophisticated integration method for the complex and rich multi-modal visual dialog task: We first exploit the local multi-modal structures through GNNs before propagating inter-modal information in a cascaded manner via hub-nodes to finally enhance the hidden states of each BERT layer.
To the best of our knowledge, we are the first to propose this for the visual dialog task.

\subsection{Language Models for Visual Dialog}
ViLBERT \cite{lu2019vilbert} and XLMERT \cite{tan2019lxmert} were among the first attempts to leverage pre-trained LMs for vision-language tasks (e.g. VQA \cite{VQA}, VCR \cite{zellers2019vcr}, and image retrieval \cite{young-etal-2014-image}).
Concretely, they used a transformer encoder as a backbone and deployed a two-stream architecture to separately encode text and visual input.
VisDial-BERT \cite{murahari2020large} built on top of ViLBERT and specifically adapted it to the visual dialog task, achieving new state-of-the-art performance on VisDial, and thus becoming the standard baseline for this dataset.
More recently, UTC \cite{Chen2022} was introduced as an improvement of VisDial-BERT and used two inter-task contrastive losses to improve training.
Another class of models used a single stream to encode the multi-modal input such as
B2T2 \cite{b2t2}, VisualBERT \cite{visualbert}, VL-BERT \cite{VL_BERT}, and UNITER \cite{chen2020uniter}.
VD-BERT \cite{Wang2020} adapted a single stream VQA model \cite{Zhou_Palangi_Zhang_Hu_Corso_Gao_2020} to the visual dialog task and managed to achieve good performance without relying on external datasets.

Contrarily to the aforementioned works, our model relies on the power of cascaded GNNs to exploit local structures within each modality and, thus produce more fine-grained representations for subsequent BERT layers.
To the best of our knowledge, this combination of GNNs and pre-trained LMs has not been explored before for visual dialog.

\section{Method}

As shown in \autoref{fig:method}, \vdgr\, consists of four main components: a) a graph construction and pruning module, (b) a backbone operating on alternating cascaded multi-modal graphs and BERT layers, (c) multiple loss heads
(next sentence prediction $\mathcal{L}_{nsp}$, masked language modelling $\mathcal{L}_{mlm}$, masked region modelling $\mathcal{L}_{mrm}$, and multi-modal graph edge masking ($\mathcal{L}^\mathcal{I}_{gem}$, $\mathcal{L}^\mathcal{Q}_{gem}$, $\mathcal{L}^\mathcal{H}_{gem}$))
, and (d) a prediction module to rank a set of candidate answers.

\subsection{Problem Formulation}
Given a  question $\texttt{Q}_\texttt{t}$ grounded on an image $\texttt{I}$ at $t$-th turn, as well as its dialog history
$\texttt{H}_\texttt{t} = \{\texttt{C}, \texttt{(Q}_\texttt{1}, \texttt{A}_\texttt{1}\texttt{)}, ..., \texttt{(Q}_{\texttt{t-1}}, \texttt{A}_{\texttt{t-1}} \texttt{)}\}$ (where $\texttt{C}$ denotes the image caption), the model is tasked to predict its answer $\texttt{A}_\texttt{t}$ by ranking a list of $N=100$ answer candidates $\{\hat{\texttt{A}}_\texttt{t}^\texttt{1}, \hat{\texttt{A}}_\texttt{t}^\texttt{2}, ..., \hat{\texttt{A}}_\texttt{t}^{\texttt{100}}\}$.

\subsection{Graph Construction and Pruning}
\noindent\textbf{Image Modality. }
The image graph constructor $\mathcal{G_I}(.)$ treats each object in the image as a node and relies on spatial relationships to construct the graph topology, i.e. the adjacency matrix, where each edge represents a relationship between two objects in the image.
The objects $\mathbf{I} = \{\mathbf{v}_1, ..., \mathbf{v}_{N_i}\}$ are obtained
using Faster R-CNN \cite{NIPS2015_14bfa6bb} pre-trained on Visual Genome \cite{krishna2017visual} where each object feature $\mathbf{v}_i$ is a $2048$ dimensional vector and $N_i = 36$.
Similar to \cite{Yao2018}, we distinguish between $11$ relations based on the $(x_1, y_1, x_2, y_2)$ object coordinates.
Specifically, the overlapping region and spatial coordinates of two regions are used to judge whether an edge exists between them or not.

\noindent\textbf{Question Modality.}
The question graph constructor $\mathcal{G_Q}(.)$ treats each word in the question as a node and relies on dependency relations to construct the graph topology, i.e. the adjacency matrix, where each edge represents a relationship between two words in the question.
To extract these relationships, we use the neural dependency parser of Stanza \cite{qi2020stanza} that yields $47$ relations.

\noindent\textbf{History Modality.}
The history graph constructor $\mathcal{G_Q}(.)$ treats each dialog round as a node and relies on coreference relations to construct the graph topology, i.e. the adjacency matrix, where each edge represents a relationship between two rounds in the history.

As can be seen in \autoref{fig:method}, the topology of all multi-modal graphs is computed \textit{once} during a pre-processing stage and is always kept constant over time.
We refer to the supplementary material for additional details.

\subsection{Proposed Layer}
\subsubsection{Transformer Features}
Inline with previous works, we use VisDial-BERT \cite{murahari2020large} layers within our novel GNN-enhanced approach.
For a given image-question pair at round $t$, we first concatenate the caption $\texttt{C}$, the previous dialog rounds $\texttt{H}_\texttt{t}$, the current question $\texttt{Q}_\texttt{t}$, and a candidate answer $ \tilde{\texttt{A}}_\texttt{t}$ to form the textual input
\begin{equation}
    \mathbf{T} =\{ \texttt{[CLS]}\, \texttt{C} \, \texttt{[SEP]} \, \texttt{Q}_\texttt{1}\texttt{[SEP]}  \texttt{A}_\texttt{1}, .., \texttt{Q}_\texttt{t} \, \texttt{[SEP]} \, \tilde{\texttt{A}}_\texttt{t}\},
\end{equation}
where \texttt{[CLS]} and \texttt{[SEP]} are the special classification and separation tokens, respectively.
As in \cite{murahari2020large,Wang2020,vd_pcr,Chen2022}, 
we use a special learnable token \texttt{[IMG]}, and initialise it using mean pooling of the object sequence.
The hidden features of the \texttt{[IMG]} token are used in conjunction with those of \texttt{[CLS]} by means of element-wise multiplication to produce the final input features of the NSP head.

Finally, the two sequences are used as input for the language and vision streams of the VisDial-BERT layer to obtain the hidden states $\mathbf{T}_h^{(l)}$ and $\mathbf{I}_h^{(l)}$, where $l$ is the \vdgr\, layer index.
\begin{table*}[!t]
    \begin{minipage}{1\linewidth}
          \begin{center}
              \scalebox{0.8}[0.8]{
              \begin{tabular}{lcccccccccccc}
                \toprule
                \multirow{2}*{\textbf{Method}} 
                & \multicolumn{6}{c}{\textbf{VisPro}} & \multicolumn{6}{c}{\textbf{VisDialConv}} \\
                \cmidrule(r){2-7} \cmidrule(r){8-13}
                 & \textbf{NDCG}$\uparrow$ & \textbf{MRR}$\uparrow$ & \textbf{R@1}$\uparrow$ & \textbf{R@5}$\uparrow$ & \textbf{R@10}$\uparrow$ & \textbf{Mean}$\downarrow$ &
                %-----------
                \textbf{NDCG}$\uparrow$ & \textbf{MRR}$\uparrow$ &\textbf{R@1}$\uparrow$ & \textbf{R@5}$\uparrow$ & \textbf{R@10}$\uparrow$ &  \textbf{Mean}$\downarrow$ \\
                %-------------------------------------------------------------------------------------------------------------------
                \midrule
                 MCA-I \cite{Agarwal2020}
                & $59.80$ & $57.88$ & $45.39$ & $72.24$ & $82.76$ & $5.84$ 
                & $52.07$ & $55.55$ & $41.65$ & $72.47$ & $83.81$ & $5.92$\\
                
                MCA-I-HConcQ \cite{Agarwal2020}
                & $61.08$ & \underline{$61.79$} & \underline{$48.95$} & \underline{$77.50$} & $86.58$ & \underline{$4.72$}
                & $54.84$ & $62.06$ & $47.42$ & \underline{$80.10$} & \underline{$88.87$} & \underline{$4.37$}\\
                
                 MCA-I-HGuidedQ \cite{Agarwal2020}
                & $61.35$ & $60.13$ & $47.11$ & $75.26$ & $86.18$ & $5.23$
                & $53.81$ & \underline{$62.29$} & \underline{$48.35$} & \underline{$80.10$} & $88.76$ & $4.42$\\
                
                MCA-I-VGH \cite{Agarwal2020}
                & $61.68$ & $59.33$ & $46.18$ & $75.53$ & \underline{$86.71$} & $5.07$
                & $55.48$ & $58.48$ & $44.54$ & $74.95$ & $86.19$ & $5.18$ \\
                
                 MCA-I-H \cite{Agarwal2020}
                & \underline{$61.72$} & $59.62$ & $45.92$ & $77.11$ & $86.45$ & $4.85$
                & $53.01$ & $61.24$ & $47.63$ & $79.07$ & $87.94$ & $4.77$\\

                Student \cite{student}
                & $-$ & $-$ & $-$ & $-$ & $-$ & $-$
                & \underline{$59.30$} & $-$ & $-$ & $-$ & $-$ & $-$\\

               \midrule

               \vdgr
               & {$\mathbf{67.09}$}    & {$\mathbf{66.82}$}  & {$\mathbf{54.47}$} & {$\mathbf{81.71}$} & {$\mathbf{91.44}$} & {$\mathbf{3.54}$}
               & {$\mathbf{60.35}$} & {$\mathbf{69.89}$}  & {$\mathbf{57.21}$} & {$\mathbf{85.97}$} & {$\mathbf{92.68}$}    & {$\mathbf{3.15}$}  \\              
              \bottomrule
              \end{tabular}
              }
          \end{center}
          \caption{Performance comparison on VisPro.
          The best and second-best results are in \textbf{bold} and \underline{underlined}, respectively.
            $\uparrow$ indicates higher is better and $\downarrow$ indicates lower is better.
          } 
          \label{tab:visdialconv}
      \end{minipage}
\end{table*}
\subsubsection{Spatial-Temporal GNN Features}
\paragraph{Node Features.}
As illustrated in \autoref{fig:method}, the multi-modal GNNs of the $l$-th \vdgr\, layer get their node features from the hidden states of the previous VisDial-BERT layer (or from the embedding layers in the first step).
It is worth noting that the special tokens \texttt{[IMG]} and \texttt{[CLS]} are not included in the graph features.
The nodes of the image and question graphs are gathered from the image and question token embeddings of the $(l-1)$-th layer, i.e. $\{\mathbf{I}_{h, i}^{(l-1)}\}_{i=1}^{N_i}$ and $\{\mathbf{T}_{h, i}^{(l-1)}\}_{i=s_q}^{s_q + N_q}$, where  $s_q$ and $N_q$ denote the index of the first question token and the length of the question, respectively.
We use the special \texttt{[SEP]} tokens to represent each dialog round in the history and gather their hidden states from $\mathbf{T}_h^{(l-1)}$ to get the node features of the history graph.
The node features of the question and history graphs have to be extracted carefully
since the textual input, i.e. where the question starts and ends and where the \texttt{[SEP]} tokens are located, varies within $\mathbf{T}_h^{(l-1)}$ for each dialog and round.

\paragraph{Hub-node Features.}
To make each modality aware of the other in a cascaded manner, we introduce hub-nodes as illustrated in \autoref{fig:method}.
As a result, the history becomes aware of the image, the question becomes aware of the history, and finally the image becomes aware of the question.
The hub-node within each graph links to all other nodes using a special edge feature to propagate information on a \textit{local scale} from one modality to the other before applying self- and cross-attention.
To obtain the hub-node features, we train attention-based graph embeddings for each modality, i.e. $f_{\theta_\mathcal{I}}(.)$, $f_{\theta_\mathcal{Q}}(.)$, and $f_{\theta_\mathcal{H}}(.)$, which take the node features of the corresponding graph and output a single vector representing it.
For example, the image hub-node
$\mathcal{I}$, is computed from the image graph node features $\mathbf{I}_{{G}}$ of the $l$-th layer as follows:
\begin{align}
    \mathcal{I} &=  f_{\theta_\mathcal{I}}(\mathbf{I}_{{G}}) = \sum_i  \alpha_i \mathbf{v}_i \, \, \mathrm{for} \,\, \mathbf{v}_i \in \mathbf{I}_{{G}}, \\
    \alpha &= \{\alpha\}_i = \mathrm{MLP}(\mathbf{I}_{{G}}),
\end{align}
where MLP is a multi-layer perceptron that maps from the node features' vector space to $\mathbb{R}$ and $\theta_\mathcal{I}$ is the set of learnable parameters of the embedding.
The question hub-node $\mathcal{Q}$ and history hub-node $\mathcal{H}$ are obtained in the same manner.

\subsubsection{Graph Aggregation}
Our multi-modal graphs are a variant of graph attention networks \cite{gat}, although other types of GNNs can be used within each \vdgr\, layer.
Each multi-modal graph uses $K$ layers (purple boxes in \autoref{fig:method}) to propagate information between the nodes.
For example, the $k$-th image graph layer receives the node features from the previous graph layer and updates them as follows:
\begin{align}
    &\mathbf{I}^{(k)}_G = \{\mathbf{v}_1^{(k)},.., \mathbf{v}_{N_i}^{(k)}, \underbrace{\mathbf{v}_{N_i + 1}^{(k)}}_{=\mathcal{Q}} \} = \mathrm{GNN}_{\mathcal{I}}(\mathbf{I}^{(k-1)}_G),\\
    &\mathbf{v}_i^{(k)} = \mathrm{GeLU}\left(\displaystyle\mathrel{\mathop{\parallel}_{h=1}^H} \tilde{\mathbf{v}}_i^{(k, h)} + \mathbf{v}_i^{(k-1)} \right),\\
    &\tilde{\mathbf{v}}_i^{(k, h)} = f(\sum_{v_s\in \mathcal{N}_{v_t}} \alpha_{s\rightarrow t}^{(k,h)} \mathbf{m}_{s\rightarrow t}^{(k,h)}) \,\, \forall 1\leq h \leq H,
\end{align}
where $\parallel$ and $H$ are the concatenation operation and the number of GNN attention heads, respectively.
$\mathcal{N}_{v_t}$ represents the neighbourhood of node $v_t$, $\alpha_{s \rightarrow t}$ denotes the attention weight that scales the message $\mathbf{m}_{s \rightarrow t}^{(k,h)}$ between a source node $\mathbf{v}_s$ and a target node $\mathbf{v}_t$, and $f$ is a linear layer.
The messages $\mathbf{m}_{s \rightarrow t}^{(k,h)}$ between the nodes are computed following:
\begin{equation}
    \mathbf{m}_{s \rightarrow t}^{(k,h)} = g_h(\mathbf{v}_s^{(k-1)},  \mathbf{e}_{s \rightarrow t}),
\end{equation}
where $g_h$ is a linear layer and $\mathbf{e}_{s \rightarrow t}$ is the edge feature between the nodes $\mathbf{v}_s$ and $\mathbf{v}_t$.
We omitted the $l$ index of the \vdgr\, layer in the previous equations for brevity.
The $k$-th question and history graph layers update their node features $\mathbf{Q}_G^{(k)}$ and $\mathbf{H}_G^{(k)}$ in the same manner.

\subsubsection{Hidden States Enhancement}
The outputs of the multi-modal graphs of the last $K$-th layer, i.e. $\mathbf{I}_G^{(l,K)}$, $\mathbf{Q}_G^{(l,K)}$ and $\mathbf{H}_G^{(l,K)}$, are used to enhance the hidden states of the following BERT layer.
First, the GNN features are scattered back to their corresponding places within the VisDial-BERT hidden states.
Then, we apply a fusion operation inspired by the idea of residual connections \cite{resnet} as illustrated in \autoref{fig:method}.
Our experiments show that this step is crucial and leads to significant improvements in performance.
These operations can be summarised as follows:
\begin{align}
    &\tilde{\mathbf{I}}_h^l = \mathbf{I}_h^l \oslash	 (\mathbf{I}_G^{(l, K)}, \mathrm{Idx}_v), \,\,\hat{\mathbf{I}}_h^l = \lambda \mathbf{I}_h + (1-\lambda) \tilde{\mathbf{I}}_h^l,\\
    &\tilde{\mathbf{T}}_h^l = \big(\mathbf{T}_h^l \oslash (\mathbf{Q}_G^{(l, K)}, \mathrm{Idx}_q)\big)  \oslash \big(\mathbf{H}_G^{(l, K)}, \mathrm{Idx}_h\big),\\
    &\hat{\mathbf{T}}_h^l = \lambda \mathbf{T}_h + (1-\lambda) \tilde{\mathbf{T}}_h^l \quad \mathrm{for} \,\, \lambda \in [0, 1], \label{eq:res}
\end{align}
where $\oslash$ denote the scatter operation and $\mathrm{Idx}_*$ the indices of the graph nodes features with respect to the BERT hidden states.
The final enhanced features $\hat{\mathbf{I}}_h^l$ and $\hat{\mathbf{T}}_h^l$ are passed to the next BERT layer.

\subsection{Loss Heads}
\label{sec:losses}
We complement the traditional losses used for the visual dialog task (masked language modelling $\mathcal{L}_{\textrm{mlm}}$, masked region modelling $\mathcal{L}_{\textrm{mrm}}$, and next sentence prediction $\mathcal{L}_{\textrm{nsp}}$) with graph edge masking $\mathcal{L}_{\textrm{gem}}$ to improve learning of the local structure of each modality, and thus to enhance the feature representation of our multi-modal graphs.

\paragraph{Masked Language and Region Modelling.}
Similar to masked language modelling introduced in \cite{bert}, we randomly masked $10\%$ of the text tokens and image objects with the special token \texttt{[MASK]} and the model had to recover them based on the surrounding tokens and cross-modal clues:
\begin{align}
    \mathcal{L}_{\textrm{mlm}} &= -\mathbb{E}_{(\mathbf{w}, \mathbf{I})\sim S_{tr}}\left[log P(w_m | \mathbf{w}_{\backslash m}, \mathbf{h}_\texttt{[IMG]})\right],\\
    \mathcal{L}_{\textrm{mrm}} &= -\mathbb{E}_{(\mathbf{w}, \mathbf{I})\sim S_{tr}}\left[log P(\mathbf{v}_m | \mathbf{I}_{\backslash m}, \mathbf{h}_\texttt{[CLS]})\right],
\end{align}
where $w_m$, $\mathbf{v}_m$, and $S_{tr}$ denote the masked words, the masked regions, and the training set, respectively.

\paragraph{Next Sentence Prediction.}
The aim of next sentence prediction (NSP) is to identify whether the appended answer $\tilde{\texttt{A}}_\texttt{t}$ is correct or not, i.e.
\begin{equation}
    \mathcal{L}_{\textrm{nsp}} = -\mathbb{E}_{(\mathbf{w}, \mathbf{I})\sim S_{tr}}\left[log P(y | \aleph(\mathbf{w}, \mathbf{I}))\right],
\end{equation}
where $y\in \{0,1\}$ is a ground-truth binary label, and $\aleph(., .)$ is the binary answer prediction head operating on the element-wise product of the \texttt{[IMG]} and \texttt{[CLS]} token representations.

\begin{figure}[!t]
    \begin{minipage}{1\linewidth}
        \centering
        \scalebox{0.99}[0.99]{
            \includegraphics[width=\textwidth]{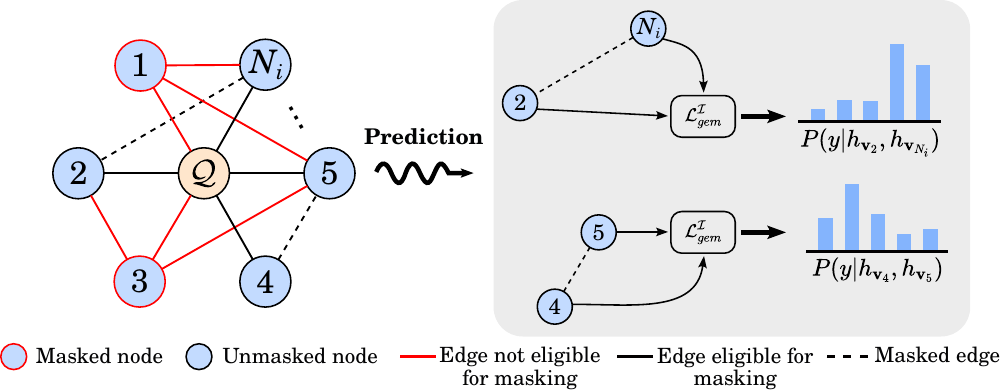}
        }
        \caption{
            Only edges connecting two unmasked node features are eligible for masking with a $15\%$ probability.
        }
        \label{fig:gem}
    \end{minipage}
\end{figure}

\paragraph{Graph Edge Masking.}
We introduce the multi-modal graph edge masking to alleviate the potential problem of having well-trained BERT layers but not sufficiently trained GNNs.
Given that, by design, our GNNs receive their features from the previous BERT layer, they inherit some masked node representations.
In order to make the edge prediction task stable, we only masked $15\%$ of the edges connecting two unmasked node features as illustrated in \autoref{fig:gem}.
The representations of these nodes were then used to predict the masked edges.
For the example of the image graph, this results in the following loss:
\begin{equation}
    \mathcal{L}^\mathcal{I}_{\textrm{gem}} = -\mathbb{E}_{(\mathbf{w}, \mathbf{I})\sim S_{tr}}\left[log P(y_m^{(i,j)} | \mathbf{h}_{\mathbf{v}_i}, \mathbf{h}_{\mathbf{v}_j} )\right],
\end{equation}
where $y_m^{(i,j)}$ is the ground-truth edge type between the nodes $\mathbf{v}_i$ and $\mathbf{v}_j$.
The question and history edge graph masking losses $\mathcal{L}^\mathcal{Q}_{\textrm{gem}}$ and $\mathcal{L}^\mathcal{H}_{\textrm{gem}}$ are obtained in a similar manner.

\paragraph{Total Loss.}
We adopt a two-stage approach to train our model.
First, we train it on a warm-up task of masked token and graph edge prediction, i.e. using the total loss $\mathcal{L}_{\textrm{warm}}$:
\begin{align}
    &\mathcal{L}_\textrm{warm} = \alpha_{1} (\mathcal{L}_{\textrm{mlm}} + \mathcal{L}_{\textrm{mrm}}) + \alpha_{2} \mathcal{L}_{\textrm{GEM}}, \\
    &\mathcal{L}_{\textrm{GEM}} = \mathcal{L}_{\textrm{GEM}}^\mathcal{I} + \mathcal{L}_{\textrm{GEM}}^\mathcal{Q} + \mathcal{L}_{\textrm{GEM}}^\mathcal{H}.
\end{align}
Then, we only train the model based on the visual dialog loss \begin{equation}
    \mathcal{L}_{\textrm{VD}} 
        = \mathcal{L}_{\textrm{mlm}} + \mathcal{L}_{\textrm{mrm}} + \mathcal{L}_{\textrm{nsp}}.
\end{equation}

\section{Experiments}
\label{sec:experiment}

\subsection{Datasets}
We evaluated \vdgr\, on the challenging VisDial v0.9 and VisDial v1.0 datasets.
VisDial v0.9 has circa $83$k training and $40$k validation dialogs.
The more recent v1.0 version consists of about $123$k, $2$k, and $8$k images for training, validation, and testing, respectively.
Each image comes with a caption and $10$ question-answer pairs; each question turn is associated with $100$ candidate answers.
The validation data and part of the training data of VisDial v1.0 provide dense annotations for the candidate answers.
Furthermore, we evaluated our model on two additional datasets, i.e. VisPro \cite{vispro} and VisDialConv~\cite{Agarwal2020}.

\subsection{Quantitative Results}
\begin{table}[!t]
    \begin{minipage}{1\linewidth}
        \begin{center}
          \scalebox{0.8}[0.8]{
          \begin{tabular}{lccccc}
            \toprule
            \textbf{Method} & \textbf{MRR}$\uparrow$ & \textbf{R@1}$\uparrow$ & \textbf{R@5}$\uparrow$ & \textbf{R@10}$\uparrow$ & \textbf{Mean}$\downarrow$ \\
            \midrule
            {MN} \cite{visdial}                   & $59.65$ & $45.55$ & $76.22$ & $85.37$ & $5.46$  \\
            {CoAtt} \cite{yu2017multi}            & $63.98$ & $50.29$ & $80.71$ & $88.81$ & $4.47$  \\
            {HCIAE} \cite{lu2017best}             & $62.22$ & $48.48$ & $78.75$ & $87.59$ & $4.81$  \\
            {CorefNMN} \cite{coref-nmn}           & $64.10$ & $50.92$ & $80.18$ & $88.81$ & $4.45$  \\
            {RvA} \cite{niu2019recursive}         & $66.34$ & $52.71$ & $82.97$ & $90.73$ & $3.93$  \\
            {Student}  \cite{student}             & $60.03$ & $50.40$ & $70.74$ & $77.15$ & $12.13$ \\
            {DVAN} \cite{ijcai2019p693}           & $66.67$ & $53.62$ & $82.85$ & $90.72$ & $3.93$  \\
            {VD-BERT} \cite{Wang2020}             & $70.04$ & $57.79$ & $85.34$ & $92.68$ & $4.04$  \\
            VisDial-BERT \cite{murahari2020large} & \underline{$71.99$} & \underline{$59.41$} & \underline{$87.92$} & \underline{$94.59$} & \underline{$2.87$} \\
            \midrule
            {\vdgr}                     & $\mathbf{74.50}$ & $\mathbf{62.10}$ & $\mathbf{90.49}$ & $\mathbf{96.37}$ & $\mathbf{2.45}$ \\
            \bottomrule
          \end{tabular}
          }
        \end{center}
        \caption{Performance comparison on the \textit{val} split of VisDial v0.9 dataset.
          NDCG is not supported in this version of the dataset.
          } 
          \label{tab:visdial_v09_val}
      \end{minipage}
\end{table}

\paragraph{VisDialConv \& VisPro.}
First, we evaluated our model on VisPro
\footnote{Same subset as in \cite{Agarwal2020}.}
and VisDialConv which were introduced to verify the role of dialog history in answering the current question  $\texttt{Q}_\texttt{t}$.
We compared \vdgr\, to the baselines introduced in \cite{Agarwal2020} as well as the most recent  
Student model \cite{student}.
As can be seen from \autoref{tab:visdialconv}, \vdgr\, significantly outperformed all MCA variants across all metrics on both datasets.
Specifically, it increased the performance of the baselines by over $5$ absolute points on NDCG and MRR on VisPro.
On VisDialConv, \vdgr\, increased the top performance by over $4$ absolute points on the same metrics.
\autoref{tab:visdialconv} also shows that our model managed to surpass the Student model by over $1$ NDCG absolute point although it was trained on circa $13$M additional images.  

\begin{table}[!t]
    \begin{minipage}{1\linewidth}
        \begin{center}

          \scalebox{0.69}[0.69]{
          \begin{tabular}{lcccccc}
            \toprule
            \textbf{Method} & \textbf{NDCG}$\uparrow$ & \textbf{MRR}$\uparrow$ & \textbf{R@1}$\uparrow$ & \textbf{R@5}$\uparrow$ & \textbf{R@10}$\uparrow$ & \textbf{Mean}$\downarrow$ \\
            \midrule
            {LTMI} \cite{Nguyen2020} & $62.72$ & $62.32$ & $48.94$ & $78.65$ & $87.88$ & $4.86$ \\
            {VD-BERT} \cite{Wang2020}& $63.22$ & $67.44$ & $54.02$ & $83.96$ & $92.33$ & $3.53$ \\
            {VisDial-BERT} \cite{murahari2020large}     & $60.96$ & $67.17$ & $53.42$ & $84.41$ & $92.62$ & $3.41$ \\
            {MCA}  \cite{Agarwal2020}            & $60.27$ & $64.33$ & $51.12$ & $80.91$ & $89.65$ & $4.24$ \\
            {UniMM-UL}  \cite{wang2022unified}       & $62.86$ & $53.49$ & $42.70$ & $65.03$ & $74.58$ & $10.65$ \\
            {UTC}  \cite{Chen2022}            & $63.22$ & $68.58$ & $55.48$ & $85.38$ & $93.20$ & $3.28$ \\
            {Student} \cite{student}          & {$\mathbf{65.47}$} & $53.19$ & $43.08$ & $64.09$ & $71.51$ & $14.34$ \\
            {VD-PCR}   \cite{vd_pcr}        & $64.16$ & \underline{${69.71}$} & \underline{${56.79}$} & \underline{${85.82}$} & \underline{${93.64}$} & \underline{${3.15}$} \\

            \midrule
            {\vdgr}  & \underline{${64.32}$} & {$\mathbf{69.91}$} & {$\mathbf{57.01}$} & {$\mathbf{86.14}$} & {$\mathbf{93.74}$} & {$\mathbf{3.13}$} \\
            \bottomrule
          \end{tabular}
          }
        \end{center}

          \caption{
          Performance comparison on the \textit{val} split of VisDial v1.0 dataset.
          } 
          \label{tab:visdial_val}
      \end{minipage}
\end{table}

\paragraph{VisDial v0.9.}
Second, we compared \vdgr\, with the state of the art on the \textit{val} split of VisDial v0.9.
As can be seen from \autoref{tab:visdial_v09_val}, our model significantly outperformed all previous models and achieved new state-of-the-art results across all metrics.
Specifically, it outperformed pre-training methods such as VisDial-BERT and VD-BERT by a large margin. 
Even more importantly, it managed to also surpass more recent models such as Student.
Specifically, \vdgr\, improved the MRR and R@1 scores by over $2.5$ absolute points compared to the second best model.

\begin{figure}[!t]
    \begin{minipage}{1\linewidth}
        \centering
        \scalebox{0.99}[0.99]{
            \includegraphics[width=\textwidth]{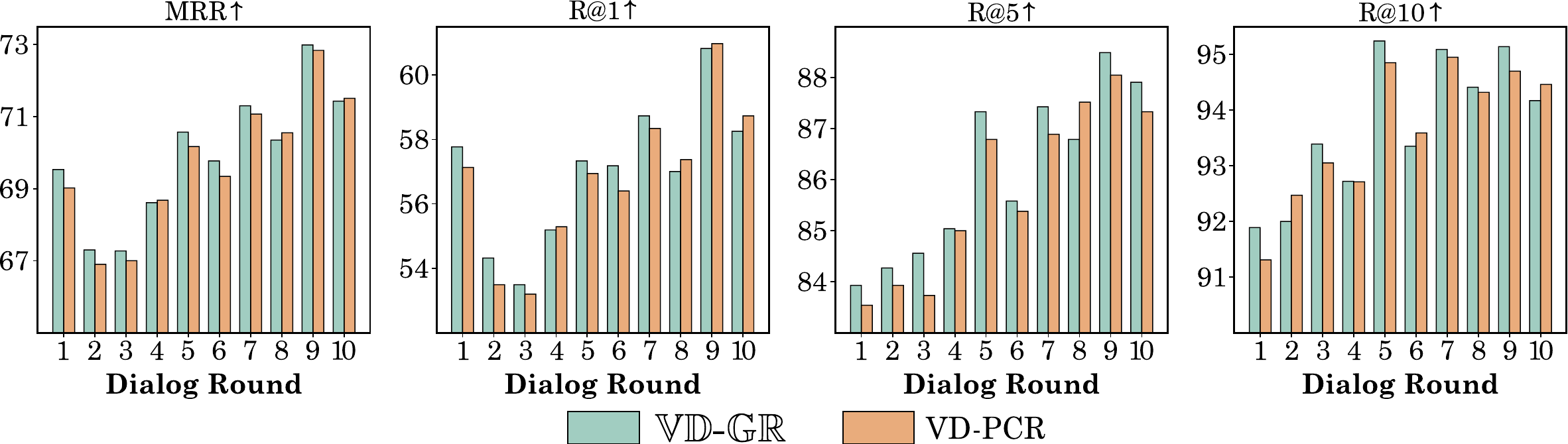}
        }
        \caption{
           Performance comparison on different dialog rounds of the VisDial v1.0 \textit{val} split.
           We only considered sparse metrics since the dense annotations used to compute the NDCG score are not defined on all 10 rounds of the validation dialogs.
        }
        \label{fig:quan_rounds}
    \end{minipage}
\end{figure}

\paragraph{VisDial v1.0 \textit{val}.}
We then compared \vdgr\, with the state of the art on the \textit{val} split of VisDial v1.0.
As can be seen from \autoref{tab:visdial_val}, our model outperformed all previous models across all five sparse metrics.
Specifically, it outperformed pre-training methods, such as VisDial-BERT and VD-BERT by a significant margin.
Even more importantly, it managed to also surpass more recent models such as UniMM-UL, VD-PCR, UTC, and Student. 
\vdgr\, improved MRR, R@1, and R@5 by over $0.2$ absolute points compared to the second best VD-PCR model.
Furthermore, we compared their performance on individual dialog rounds using the sparse metrics (MRR, R@1, R@5, and R@10).
As can be seen from \autoref{fig:quan_rounds}, \vdgr\, managed to outperform VD-PCR on almost all rounds of the dataset.

\paragraph{VisDial v1.0 \textit{test-std}.}
Finally, we compared our model with state-of-the-art published baselines on the \textit{test-std} split of the VisDial v1.0 dataset.

$\bullet$ \textit{State-of-the-art Results on Sparse Metrics:}
As can be seen from the first section of \autoref{tab:visdial_test}, \vdgr\, lifted the state-of-the-art R@5, R@10, and Mean scores from $85.38$, $93.53$, and $3.21$ achieved by VD-PCR to $85.58$, $93.85$, and $3.20$, respectively.
On the remaining metrics, our models performed on par with the state of the art.
Specifically, it reached respective scores of $68.65$ and $55.33$ on MRR and R@1, only third to UTC and VD-PCR.

$\bullet$ \textit{Fine-tuning on Dense Annotations:}
As in previous works, we fine-tuned our model on the dense annotations released by \cite{murahari2020large} in order to improve the NDCG score.
As illustrated in the second section of \autoref{tab:visdial_test}, the NDCG score sharply increased from $63.49$ to $75.95$, outperforming all previous models in the single-model setting.
However, by fine-tuning on dense annotations, we decreased the performance on the sparse metrics (MRR, R@1, R@5, R@10, and Mean).
This well-known phenomenon of the dataset is due to the misalignment of the dense and sparse annotations as previously observed by \cite{murahari2020large, Wang2020}.
In contrast to other baselines,
\vdgr\, managed to keep relatively high sparse scores (4/5 metrics are the new state of the art) after fine-tuning.

\begin{figure}[!t]
    \begin{minipage}{1\linewidth}
        \centering
        \scalebox{0.975}[0.975]{
            \includegraphics[width=\textwidth]{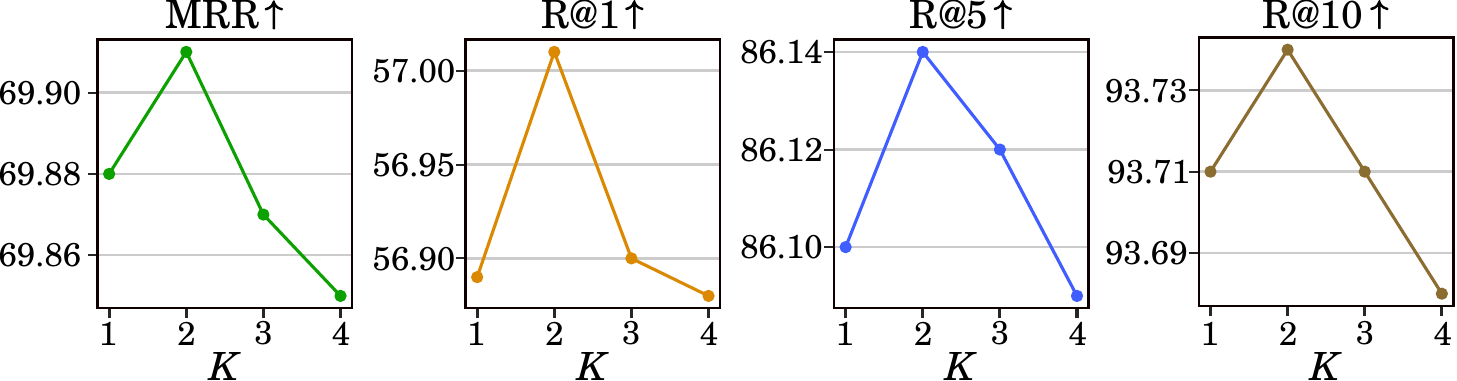}
        }
        \caption{
            Performance comparison with different number of GNN layers $K$ on the \textit{val} split of VisDial v1.0.
            Results are shown for the \textit{val} split of VisDial v1.0.
        }
        \label{fig:ablation_k}
    \end{minipage}
\end{figure}
$\bullet$ \textit{Ensemble Setting:}
As it is common practice, we fine-tuned an ensemble comprised of eight pre-trained \vdgr\, models to further improve the NDCG score.
One set of four models was fine-tuned with Cross Entropy (CE) and with a varying number for GNN layers $K$, i.e. $K=\{1,2,3,4\}$.
Each model of the second set was trained using the ListNet \cite{listnet} ranking optimisation method.
As can be seen in the last section of \autoref{tab:visdial_test}, our ensemble model reached an NDCG score of $76.43$ outperforming the closest competitor UniMM-UL with only 76.17 and, thus setting a new state of the art in the ensemble setting.
\begin{table}[!t]
    \begin{minipage}{1\linewidth}
        \begin{center}

          \scalebox{0.67}[0.67]{
          \begin{tabular}{lcccccc}
            \toprule
            \textbf{Method} & \textbf{NDCG}$\uparrow$ & \textbf{MRR}$\uparrow$ &\textbf{R@1}$\uparrow$ & \textbf{R@5}$\uparrow$ & \textbf{R@10}$\uparrow$ & \textbf{Mean}$\downarrow$ \\
            \midrule
            {MN} \cite{visdial}         & $47.50$ & $55.49$ & $46.98$ & $72.30$ & $83.30$ & $5.92$ \\
            {CorefNMN} \cite{coref-nmn}  & $54.70$ & $61.50$ & $47.55$ & $78.10$ & $88.80$ & $4.40$ \\
            {FGA} \cite{Schwartz2019}        & $56.90$ & $66.20$ & $52.75$ & $82.92$ & $91.07$ & $3.80$ \\
            {DAN} \cite{kang2019dual}       & $57.59$ & $63.20$ & $49.63$ & $79.75$ & $89.35$ & $4.30$ \\
            {LTMI} \cite{Nguyen2020}       & $59.03$ & $64.08$ & $50.20$ & $80.68$ & $90.35$ & $4.05$ \\
            {CAG} \cite{Guo2020}       & $56.64$ & $63.49$ & $49.85$ & $80.63$ & $90.15$ & $4.11$ \\
            {GOG}  \cite{gog}       & $61.04$ & $63.52$ & $50.01$ & $80.13$ & $89.28$ & $4.31$ \\
            VD-BERT \cite{Wang2020}     & $59.96$ & $65.44$ & $51.63$ & $82.23$ & $90.68$ & $3.90$ \\
            VisDial-BERT \cite{murahari2020large} & $63.87$ & $67.50$ & $53.85$ & $84.68$ & $93.25$ & $3.32$ \\
            {UTC}  \cite{Chen2022}      &  \underline{${64.60}$} & \underline{${68.70}$}  & {$\mathbf{55.73}$} & $84.93$ & $93.08$ & $3.32$ \\
            VD-PCR   \cite{vd_pcr}    & $63.55$ &  {$\mathbf{68.73}$}  &  \underline{${55.45}$} &  \underline{${85.38}$} &  \underline{${93.53}$} &  \underline{${3.21}$} \\
            {UniMM-UL} \cite{wang2022unified}  & $63.90$ & $68.14$  & $54.57$ & $85.15$ & $93.13$ & $3.27$ \\
            {$^\ddag$Student} \cite{student}   & {$\mathbf{64.91}$} & $68.44$ & $55.05$ & $85.18$ & $93.35$ & $3.23$ \\

            \midrule
            {\vdgr}      & $63.49$     & $68.65$   & $55.33$     & {$\mathbf{85.58}$}     & {$\mathbf{93.85}$}     & {$\mathbf{3.20}$} \\
            \midrule
            \midrule
            {$^{\clubsuit}$MCA} \cite{Agarwal2020}  & $72.47$ & $37.68$ & $20.67$ & $56.67$ & $72.12$ & $8.89$ \\
            {$^{\clubsuit}$VD-BERT} \cite{Wang2020}  & $74.54$ & $50.74$ & $33.15$ & $61.58$ & $77.15$ & $7.18$ \\
            {$^{\clubsuit}$VisDial-BERT} \cite{murahari2020large} & $74.47$ & $50.74$ & $37.95$ & $64.13$ & $80.00$ & $6.28$ \\
            {$^{\clubsuit}$UTC} \cite{Chen2022} & $74.32$ & $50.24$ & $37.12$ & $63.98$ & $79.88$ & $6.48$ \\
            {$^{\clubsuit}$VD-PCR} \cite{vd_pcr} & $75.30$ & $56.17$ & \underline{${45.32}$} & $68.05$ & $80.98$ & $6.15$ \\

            {$^{\clubsuit}$UniMM-UL} \cite{wang2022unified} &  \underline{${75.92}$} & \underline{${56.18}$} & $43.70$ &  \underline{${71.03}$} & {$\mathbf{84.80}$} &  \underline{${5.42}$} \\

            \midrule
            {$^{\clubsuit}$\vdgr}                    & {$\mathbf{75.95}$}  & {$\mathbf{58.30}$} & {$\mathbf{46.55}$}     & {$\mathbf{71.45}$}     &\underline{$84.52$}          & {$\mathbf{5.32}$} \\
            \midrule
            \midrule
            {$^{\diamondsuit\clubsuit}$P1+P2} \cite{qi2020two} & $74.91$ & $49.13$ & $36.68$ & $62.96$ & $78.55$ & $7.03$ \\
            {$^{\diamondsuit\clubsuit}$VD-BERT} \cite{Wang2020} & $75.35$ & $51.17$ & $38.90$ & $62.82$ & $77.98$ & $6.69$ \\
            {$^{\diamondsuit\clubsuit}$VD-PCR} \cite{vd_pcr} & $76.14$ & $56.05$ &  \underline{${44.75}$} & \underline{${68.40}$} &\underline{${82.75}$} & \underline{${5.72}$} \\
            {$^{\diamondsuit\clubsuit}$UniMM-UL} \cite{wang2022unified}  & \underline{${76.17}$} &  {$\mathbf{56.42}$} & $44.32$ &  {$\mathbf{70.23}$} &  {$\mathbf{84.52}$} & {$\mathbf{5.47}$} \\
            \midrule
            {$^{\diamondsuit\clubsuit}$\vdgr}                   & {$\mathbf{76.43}$}     & \underline{${56.35}$}   & {$\mathbf{45.18}$}     & $68.13$     & $82.18$     & $5.79$ \\
            \bottomrule

          \end{tabular}
          }
        \end{center}
          \caption{Performance comparison on the \textit{test-std} split of VisDial v1.0 dataset.
          $\clubsuit$ indicates fine-tuning on dense annotations and $\diamondsuit$ denotes ensemble model.
          $\ddag$ denotes the use of extra large datasets for training. 
          }  
          \label{tab:visdial_test}
      \end{minipage}
\end{table}

\begin{table}[!t]
    \begin{minipage}{1\linewidth}
        \begin{center}

          \scalebox{0.65}[0.65]{
          \begin{tabular}{lcccccc}
            \toprule
            \textbf{Methods} & \textbf{NDCG}$\uparrow$ & \textbf{MRR}$\uparrow$ & \textbf{R@1}$\uparrow$ & \textbf{R@5}$\uparrow$ & \textbf{R@10}$\uparrow$ & \textbf{Mean}$\downarrow$ \\
            \midrule
            \vdgr\, w/ $\lambda = 0$                 & $56.69$                 &                  $67.71$&                 $54.07$ & $85.03$                 & $92.84$                 & $3.33$ \\
            \vdgr\, w/o $\mathcal{L}_\mathrm{warm}$  & $63.76$             &             $69.83$ &             $56.84$ & \underline{${86.05}$} & \underline{${93.70}$}             & $3.15$ \\
            \vdgr\,w/o sharing                       & \underline{${64.15}$} &             $69.79$ &             $56.73$ & $86.02$             & $93.68$             & $3.15$ \\
            \vdgr\,w/o HN                            & $64.11$             & \underline{${69.86}$} & \underline{${56.88}$} & ${85.97}$           & ${93.67}$    & \underline{${3.14}$} \\
            \midrule
            \vdgr\, (Full)                                & $\mathbf{64.32}$    &    $\mathbf{69.91}$ & $\mathbf{57.01}$    & $\mathbf{86.14}$    & $\mathbf{93.74}$ & $\mathbf{3.13}$ \\
            \bottomrule
          \end{tabular}
          }
        \end{center}        
          \caption{Performance comparison of ablated versions of our model on the \textit{val} split of VisDial v1.0.
          }
          \label{tab:visdial_ablations}
      \end{minipage}
\end{table}

\begin{figure*}[!t]
    \begin{minipage}{1\linewidth}
        \centering
        \scalebox{0.99}[0.99]{
            \includegraphics[width=\textwidth]{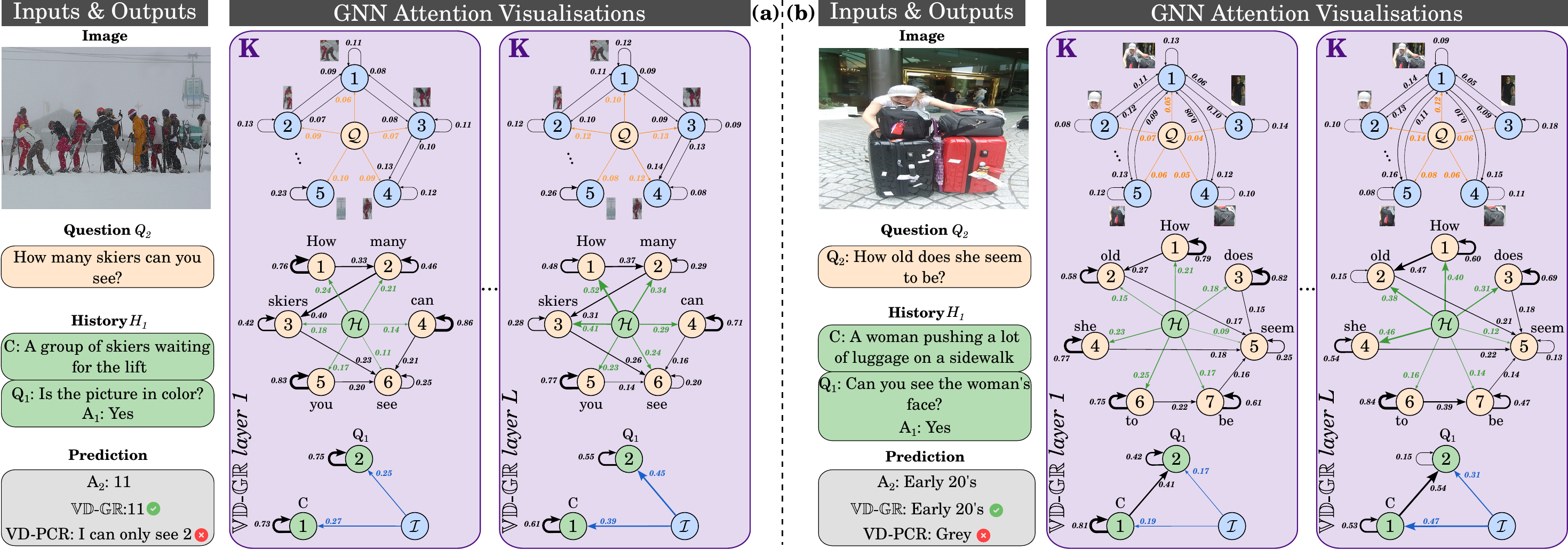}
        }
        \caption{
           Qualitative samples from the VisDial v1.0 \textit{val} split.
           The attention weights correspond to the $K$-th GNN of the first and last \vdgr\, layers. We first average them across all $H$ attention heads before re-normalising them for each node using a \texttt{softmax} function.
           For clarity, we only show a subset of the image graph nodes.
        }
        \label{fig:qualitative}
    \end{minipage}
\end{figure*}

\subsection{Ablation Study}
\paragraph{Number of GNN layers $K$.}
This is an important hyper-parameter of our model:
If $K$ is too small, then the expressive power of the GNNs will be hampered.
Contrarily, if $K$ is too large, the GNNs will suffer from over-smoothing \cite{Li_2019_ICCV}.
To this end, we increased $K$ incrementally from one to four and used the sparse metrics of the task (MRR, R@1, R@5, R@10) for validation.
As illustrated in \autoref{fig:ablation_k}, the performance of our model peaked at $K=2$ for all four metrics.
Thus, we kept this value fixed in all previous experiments unless explicitly stated otherwise.

\paragraph{Model Ablations.}
In addition to the full model, we evaluated the following ablated versions:\\
$\bullet$ \vdgr\, \textbf{w/} $\mathbf{\lambda = 0}$: This variant did not apply the proposed residual connection 
of \autoref{eq:res} while augmenting the BERT hidden states.\\
$\bullet$ \vdgr\, \textbf{w/o} $\mathbf{\mathcal{L}_\textrm{warm}}$: This variant was not trained on the warm-up task of edge prediction as discussed in Sec. \ref{sec:losses}.\\
$\bullet$\vdgr\,\textbf{w/o sharing}: This variant did not share the weights of the GNNs in each layer of our model.\\
$\bullet$ \vdgr\,\textbf{w/o HN}: This variant did not use hub-nodes to propagate the information between the multi-modal GNNs.

As can be seen from \autoref{tab:visdial_ablations}, the residual connections are essential for high performance.
Without them, \vdgr\, achieved the lowest performance across all metrics.
The same applies to the warm-up training stage: Although the performance of this ablated version improved over the previous one, it still performed significantly worse than our best model.
The results also underline the importance of sharing the GNN weights within the different \vdgr\, layers:
Although this version has more weights, it still performed worse than our best model on all metrics.
This finding was to be expected given that the local structure of each modality does not change from one \vdgr\, layer to another.
Finally, the importance of the inter-modal feature propagation using hub-nodes is highlighted by the two last rows of~\autoref{tab:visdial_ablations}: The hub-nodes enabled our model to achieve the best performance across all metrics of the \textit{val} split of the VisDial v1.0 dataset. 

\subsection{Qualitative Results}
Finally, in \autoref{fig:qualitative} we show selected qualitative samples (with more in the supplementary material) from the \textit{val} split of VisDial v1.0 alongside the ground truth answers, as well as the top-$1$ predictions of our \vdgr\, model and VD-PCR for comparison since it achieved the second best results on this split.
We make two interesting observations:
(1) Our model deals better with questions that require exploiting local structure within modalities.
For example, it managed to correctly answer $\texttt{Q}_\texttt{2} = \texttt{How many skiers can you see?}$ in the first dialog sample (see \autoref{fig:qualitative}\textcolor{red}{a}) by predicting $\texttt{11}$ whereas VD-PCR predicted \texttt{I can only see two}.
We hypothesise that this is due to the fact that our model exploits the spatial structure of the visual input more effectively using the image graph compared to VD-PCR, although the latter has access to the same visual features.
(2) \vdgr\, has more accurate semantic understanding of the question.
This is highlighted in answering $\texttt{Q}_\texttt{2} = \texttt{How old does she seem to be?}$ (referring to the woman) in the second dialog sample (see \autoref{fig:qualitative}\textcolor{red}{b}).
Whereas our model correctly predicted \texttt{Early 20's}, VD-PCR failed by answering \texttt{Grey}, which is not a semantically-sound response.
We posit that this advantage of our model is related to the fine-grained features of the question graph.

\section{Conclusion}
In this work we proposed \vdgr\, -- a novel visual dialog model that combines pre-trained language models and GNNs.
Specifically, \vdgr\, alternates between multi-modal graphs and BERT layers, and augments the hidden states of the latter with the fine-grained features obtained by the former.
\vdgr\, propagates information from one modality graph to another in a cascaded manner using hub-nodes that link to all other nodes within each modality, thereby effectively alleviating the lack of inter-modal context.
Extensive analyses underlined its effectiveness, while experiments on four challenging visual dialog datasets (VisDial v1.0, VisDial v0.9, VisDialConv, and VisPro) demonstrated its superior performance over existing methods.

\section*{Acknowledgment}
A. Bulling was funded by the European Research Council (ERC; grant agreement 801708) and L. Shi was funded by the Deutsche Forschungsgemeinschaft (DFG, German Research Foundation) under Germany’s Excellence Strategy - EXC 2075–390740016.

\appendix
\section*{Appendix}
\section{Limitations}
Although our model managed to outperform previous models on four challenging datasets, it is important to acknowledge some of its limitations:
First, \vdgr\, leverages extra data in the form of adjacency matrices of the multi-modal GNNs and relies on external models to acquire them.
Although inferring these models on the VisDial data is cheap, this approach can lead to inaccurate predictions of adjacency matrices, especially for the question and history modalities.
Thus, by keeping the graph structures constant, our model's performance might be limited by this introduced noise.
This could be remedied in future work by jointly learning the graphs' parameters as well as refining their structures over time \cite{franceschi2019learning,Chen2020,Elinas2020}.
Second, similar to almost all previous methods on this task, we did not manage to achieve new state-of-the-art performance across all metrics of this challenging dataset (see different sections of \autoref{tab:visdial_test}).
Finally, inline with previous works \cite{murahari2020large, Wang2020, kang2020reasoning, Chen2022, vd_pcr, wang2022unified}, fine-tuning our model (both in the single model as well as the ensemble setting) on dense annotations improved the most relevant metric of the dataset, i.e. the NDCG score, at the expense of the other (sparse) ones.
Although our model's performance dropped with respect to the sparse metrics, we managed to outperform previous works by achieving an NDCG score of 76.43, which is the main objective of dense annotation fine-tuning.

\section{Graph Construction and Pruning}
\begin{figure}[!h]
    \begin{minipage}{1\linewidth}
        \centering
        \scalebox{0.97}[0.97]{
            \includegraphics[width=\textwidth]{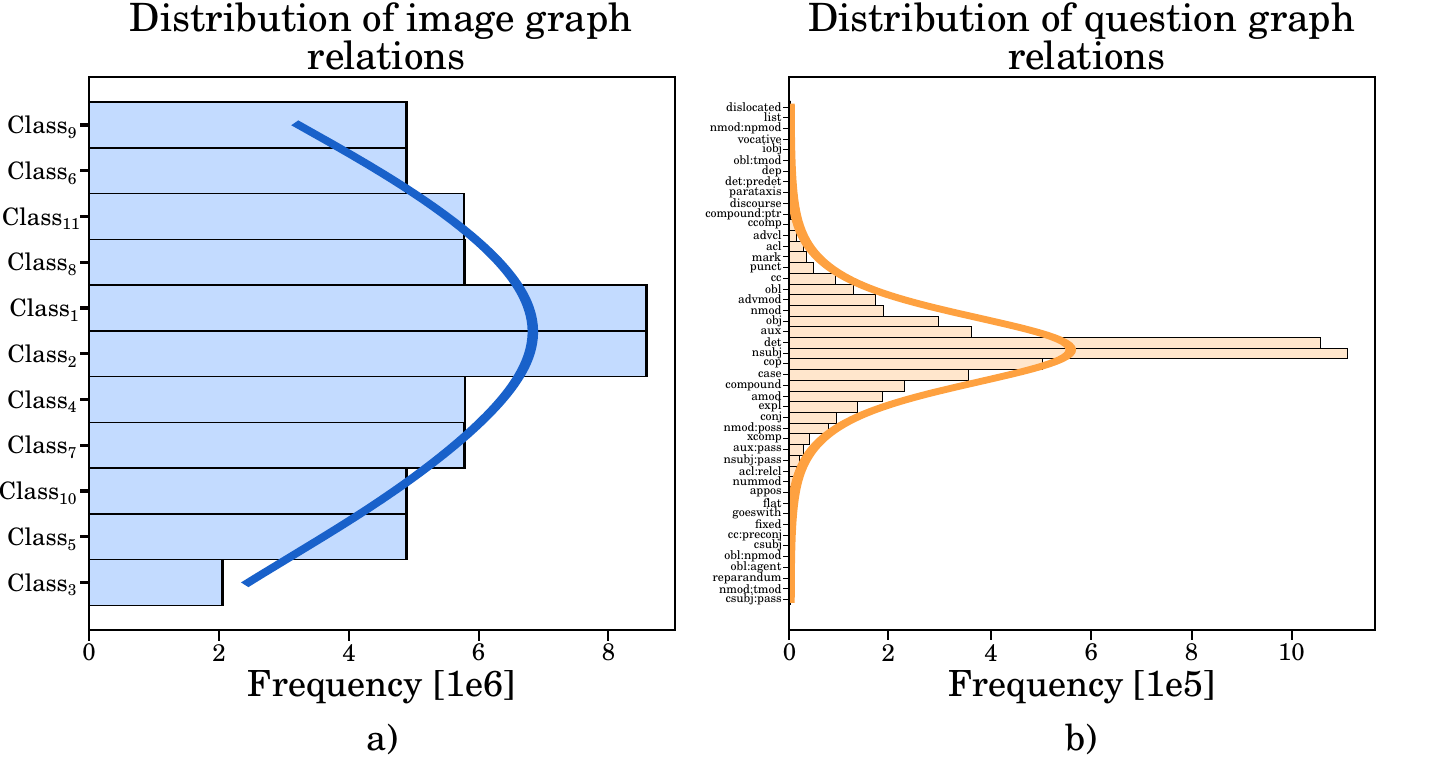}
        }
        \caption{
           \textbf{Right:} The distribution of the image graph relationship classes within the training split of VisDial v1.0.
           \textbf{Left:} The distribution of the question graph relationship classes within the training split of VisDial v1.0.
        }
        \label{fig:dist}
    \end{minipage}
\end{figure}

\begin{figure*}[!t]
    \begin{minipage}{1\linewidth}
        \centering
        \scalebox{0.97}[0.97]{
            \includegraphics[width=\textwidth]{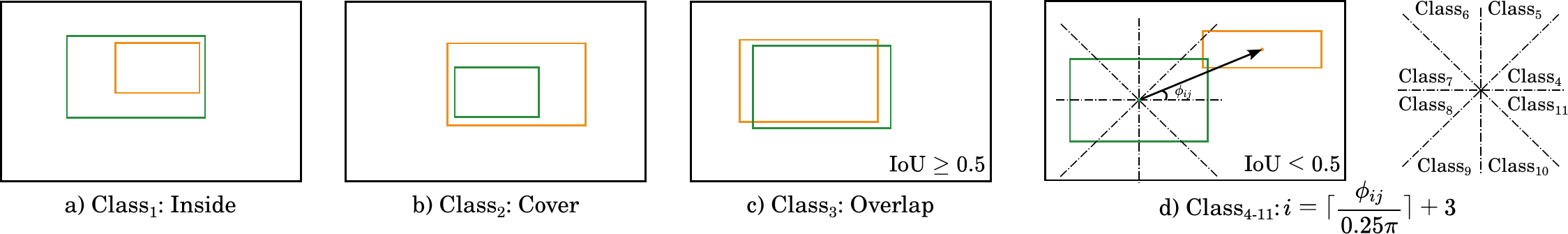}
        }
        \caption{
           The different spatial relationships (without the hub-node relationship) used in constructing the image graph.
           The orange and green rectangles correspond to the bonding boxes of two objects within the scene.
        }
        \label{fig:image_rels}
    \end{minipage}
\end{figure*}

\paragraph{Image Modality.}
Given two object features $\mathbf{v}_i$ and $\mathbf{v}_j$, their bounding boxes and centre coordinates $(x_i, y_i)$ and $(x_j, y_j)$, we computed the value of their intersection over unions $\mathrm{IoU}_{ij}$ and relative angle $\phi_{ij}$.
As shown in \autoref{fig:image_rels}, there are two spacial cases: The first occurs when the bounding box of $\mathbf{v}_i$ completely includes the bounding box of $\mathbf{v}_j$ and this class is denoted as \textit{inside} with index $i = 1$.
The second occurs when the bounding box of $\mathbf{v}_i$ is entirely covered by the bounding box of $\mathbf{v}_j$. This class is denoted as \textit{cover} with index $i=2$.
The remaining classes are solely determined by the value of $\mathrm{IoU}_{ij}$.
If $\mathrm{IoU}_{ij} \geq 0.5$, then the relationship between the objects is denoted as \textit{overlap} and has the index $i=3$.
Finally, if $\mathrm{IoU}_{ij} < 0.5$, the class index is computed as  
$$i = \lceil \frac{\phi_{ij}}{0.25\pi} \rceil+ 3.$$
By construction, all classes of index $i \neq 3$ are pairwise symmetric as can be seen from \autoref{fig:dist}\textcolor{red}{a} where we plotted the distribution of the different image graph relationship classes over the training split of VisDial v1.0. 

\begin{figure}[!h]
    \begin{minipage}{1\linewidth}
        \centering
        \scalebox{0.97}[0.97]{
            \includegraphics[width=\textwidth]{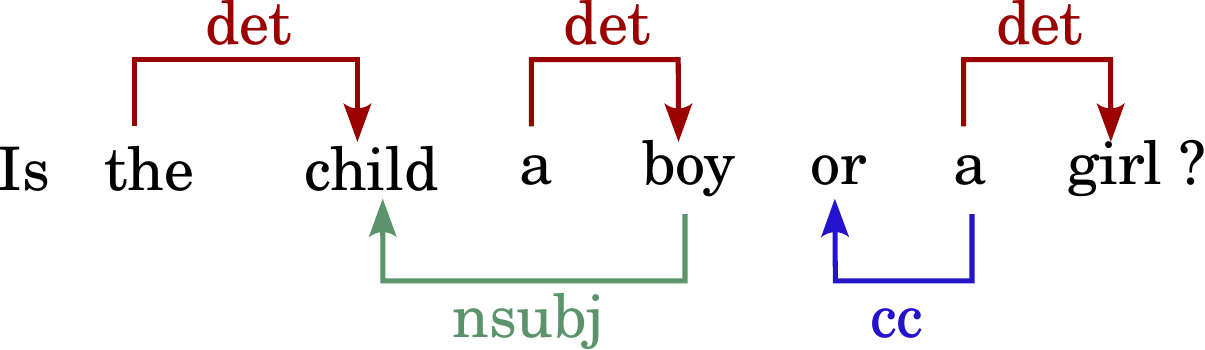}
        }
        \caption{
           A qualitative sample of the dependency relationships between question word pairs.
        }
        \label{fig:dep_pars_ex}
    \end{minipage}
\end{figure}

\paragraph{Question Modality.}
The question graph relationship classes were determined by the dependency between the question words.
To this end, we input each question to the Stanza dependency parser that output the classes between the different word pairs resulting in a total of $47$ classes.
As shown in \autoref{fig:dist}\textcolor{red}{b}, the distribution of these classes within the VisDial v1.0 training split is not uniform with \texttt{det} and \texttt{nsubj} being the most frequent.
We illustrate a qualitative sample in \autoref{fig:dep_pars_ex}.

\begin{figure}[!h]
    \begin{minipage}{1\linewidth}
        \centering
        \scalebox{0.97}[0.97]{
            \includegraphics[width=\textwidth]{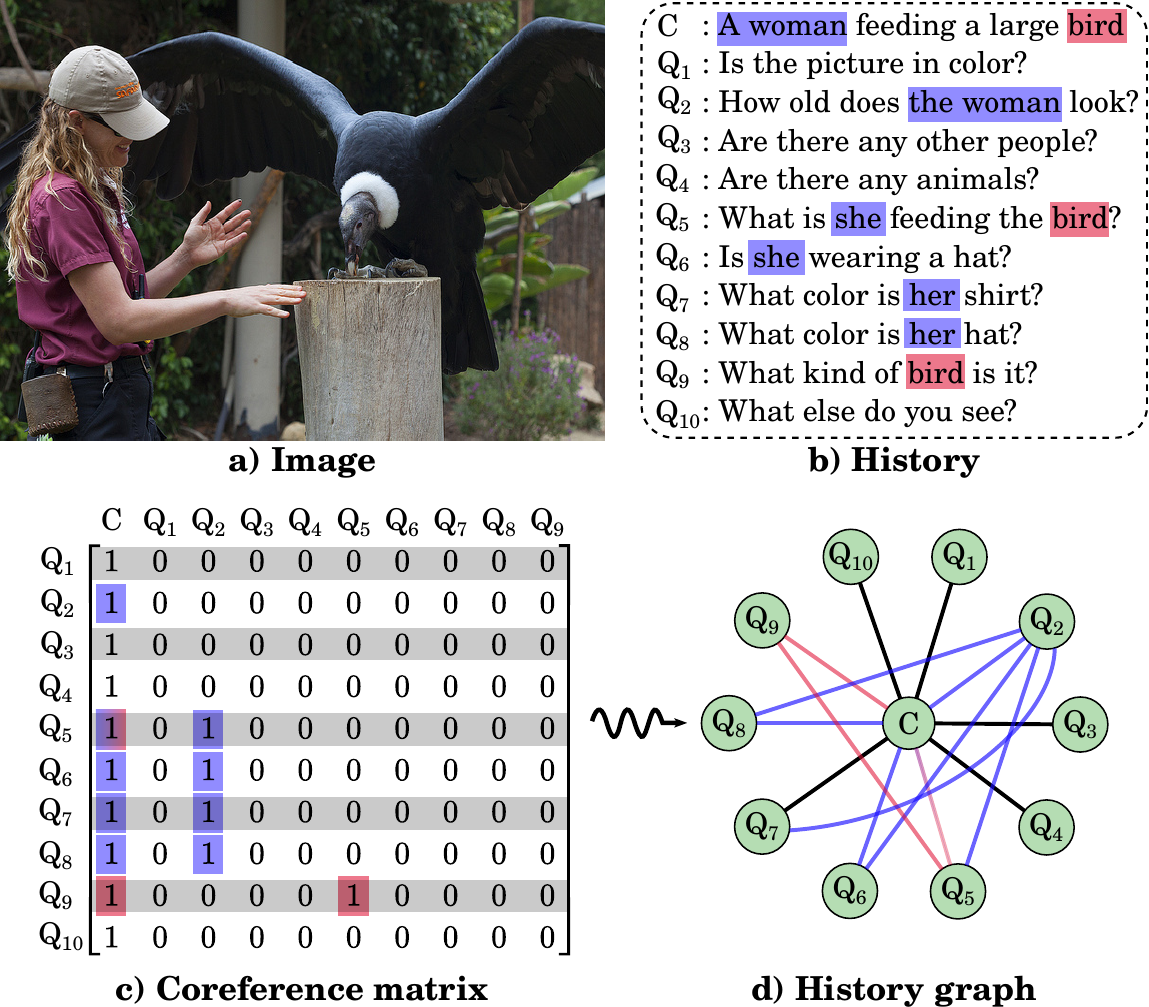}
        }
        \caption{
           A qualitative sample of the coreference relationships between different dialog rounds.
           The hub-node was not visualised for clarity.
        }
        \label{fig:coref_ex}
    \end{minipage}
\end{figure}

\paragraph{History Modality.}
We relied in coreference resolution to construct the history graph.
Specifically, an edge exists between two rounds \texttt{i} and \texttt{j} (\texttt{i} $>$ \texttt{j}) if and only if a word in round \texttt{j} was used to reference another word in round \texttt{i}.
The only exception is the caption $\texttt{C}$ that links to all upcoming rounds in the history even if there is no explicit coreference between them.
We posit that the caption is complementary to the visual input and helps the model better understand the scene.
We illustrate a qualitative sample in \autoref{fig:coref_ex}.

\begin{table}[!t]
    \begin{minipage}{1\linewidth}
        \begin{center}

          \scalebox{0.67}[0.67]{
          \begin{tabular}{lcccc}
            \toprule

            \textbf{Method} & VisDialBERT \cite{murahari2020large} & VD-BERT \cite{Wang2020} & VD-PCR \cite{vd_pcr}& \vdgr \\
            \midrule
            \# Parameters & 250M & 250M & 255M & 260M \\
            Tr. time / epoch & 0.6h & 0.6h & 1.00h & 1.05h \\
            \bottomrule
          \end{tabular}
          }
        \end{center}        
          \caption{Model complexity and runtime comparison with respect to VisDial v1.0 on our hardware setup.
          }
          \label{tab:model_complexity}
      \end{minipage}
\end{table}

\section{Model Complexity}
 The overhead for constructing the multi-modal graphs only incurs once during a \textit{cheap} offline pre-processing stage and therefore does not lead to crucial increase in compute complexity, i.e. number of trainable parameters and epoch training time, compared with previous seminal models, e.g. VisDial-BERT~\cite{murahari2020large}, VD-BERT~\cite{Wang2020}, and VD-PCR~\cite{vd_pcr}, as can be seen in \autoref{tab:model_complexity}.

\begin{table*}[!h]
    \begin{minipage}{1\linewidth}
        \begin{center}

          \scalebox{0.9}[0.9]{
          \begin{tabular}{lccc}
            \toprule
            \multirow{2}*{\textbf{Hyper-parameter}} & \multicolumn{3}{c}{\textbf{Training Stage}} \\
            \cmidrule(r){2-4}
            & \textbf{Warm-up} & \textbf{Sparse fine-tuning} & \textbf{Dense fine-tuning}\\
            \midrule
            Number of GNN layers $K$                      & $2$ & $2$ & $2$ \\
            Number of GNN heads $H$                       & $4$ & $4$ & $4$ \\
            Residual connection coefficient $\lambda$     & $0.5$ & $0.5$ & $0.5$ \\
            Dimension of $\textrm{GNN}_{\mathcal{I}}$ node features   & $1024$ & $1024$ & $1024$ \\
            Dimension of $\textrm{GNN}_{\mathcal{Q}}$ node features  & $768$ & $768$ & $768$ \\
            Dimension of $\textrm{GNN}_{\mathcal{H}}$ node features  & $768$ & $768$ & $768$ \\
            Dimension of $\textrm{GNN}_{\mathcal{I}}$ edge features  & $12$ & $12$ & $12$ \\
            Dimension of $\textrm{GNN}_{\mathcal{Q}}$ edge features  & $48$ & $48$ & $48$ \\
            Dimension of $\textrm{GNN}_{\mathcal{H}}$ edge features  & $2$ & $2$ & $2$ \\
            Dimension of $\mathrm{Linear}_{\mathcal{I}\rightarrow \mathcal{H}}(.)$ & $(1024, 768)$ & $(1024, 768)$ & $(1024, 768)$ \\
            Dimension of $\mathrm{Linear}_{\mathcal{Q}\rightarrow \mathcal{I}}(.)$ & $(768, 1024)$ & $(768, 1024)$ & $(768, 1024)$ \\
            \midrule
            Maximum number of text tokens & $256$ & $256$ & $256$ \\
            Maximum number of image regions & $37$ & $37$ & $37$ \\
            Text token mask probability & $0.1$ & $0.1$ & $-$ \\
            Image region mask probability & $0.1$ & $0.1$ & $-$ \\
            Graph edge mask probability & $0.15$ & $-$ & $-$ \\
            \midrule
            Optimiser & \texttt{Adam} & \texttt{Adam} & \texttt{Adam}\\
            Minimum learning rate of BERT parameters & $0$ & $0$ & $1\times 10^{-5}$ \\
            Minimum learning rate of GNN parameters & $0$ & $0$ & $1\times 10^{-5}$ \\
            Maximum learning rate of BERT parameters  & $5\times 10^{-6}$ & $5\times 10^{-6}$ & $2\times 10^{-5}$ \\
            Maximum learning rate of GNN parameters & $5\times 10^{-4}$ & $5\times 10^{-4}$ & $1\times 10^{-4}$ \\
            Learning rate warm-up of BERT parameters & \texttt{True} & \texttt{True} & \texttt{True} \\
            Learning rate warm-up of GNN parameters & \texttt{True} & \texttt{True} & \texttt{True} \\
            Learning rate schedule of BERT parameters & \texttt{Linear} & \texttt{Linear} & \texttt{Linear} \\
            Learning rate schedule of GNN parameters & \texttt{Linear} & \texttt{Linear} & \texttt{Linear} \\
            Training Loss & $\mathcal{L}_{\mathrm{warm}}$ & $\mathcal{L}_{\mathrm{VD}}$ & $\mathcal{L}_{\mathrm{CE}}$ / $\mathcal{L}_{\mathrm{ListNet}}$\\
            Number of epochs & $5$ & $20$ & $3$\\
            Effective batch size & $64$ & $64$ & $100$ \\
            \midrule
            GPU Model & \texttt{Tesla V100-32GB} & \texttt{Tesla V100-32GB} & \texttt{Tesla V100-32GB} \\
            Number of GPUs & $8$ & $8$ & $8$ \\
            Distributed training  & \texttt{Apex} & \texttt{Apex} & \texttt{PyTorch DP} \\
            \bottomrule
          \end{tabular}
          }
        \end{center}
        \caption{Hyper-parameter settings of \vdgr\, for the different stages of training. $\mathrm{Linear}_{\mathcal{I}\rightarrow \mathcal{H}}(.)$ and $\mathrm{Linear}_{\mathcal{Q}\rightarrow \mathcal{I}}(.)$ denote the linear layers that produce the history and image hub-node features, respectively.} 
          \label{tab:hp}
      \end{minipage}
\end{table*}
\section{Training Details}
We implemented our model using PyTorch \cite{pytorch} and trained it on a server with 8 NVIDIA Tesla V100 GPUs using an effective batch size of $64$ and Adam optimiser \cite{adam} with a linear decay learning rate schedule with warm-up.
We set the initial learning rates of the BERT and GNN weights to $5\times 10^{-6}$ and $5\times 10^{-4}$, respectively.
Furthermore, we set the loss coefficients $\alpha_1 = \alpha_2 = 1$ and the residual connection coefficient $\lambda = 0.5$.
We refer to \autoref{tab:hp} for a complete overview of our experimental setup.

\section{Additional Qualitative Results}
\begin{figure*}[t]
    \begin{minipage}{1\linewidth}
        \centering
        \scalebox{0.95}[0.95]{
            \includegraphics[width=\textwidth]{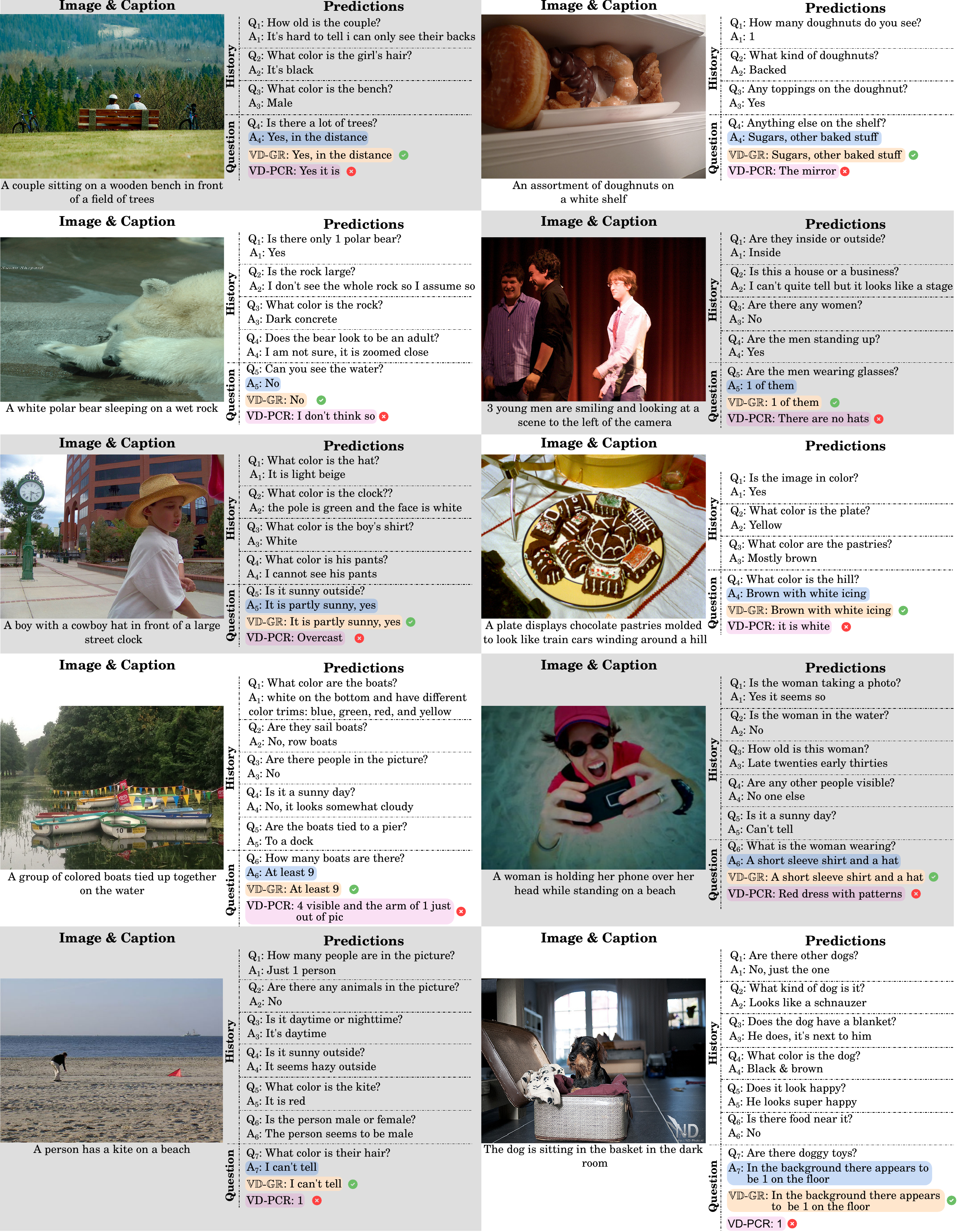}
        }
        \caption{
           Qualitative samples from the \textit{val} split of the VisDial v1.0 dataset.
           We compared the predictions of \vdgr\, (orange) with the ground truth answers (blue) and the predictions of VD-PCR (pink).
           The answers of both models correspond to the top-$1$ predictions.
        }
        \label{fig:qualitative_1}
    \end{minipage}
\end{figure*}

\begin{figure*}[t]
    \begin{minipage}{1\linewidth}
        \centering
        \scalebox{0.95}[0.95]{
            \includegraphics[width=\textwidth]{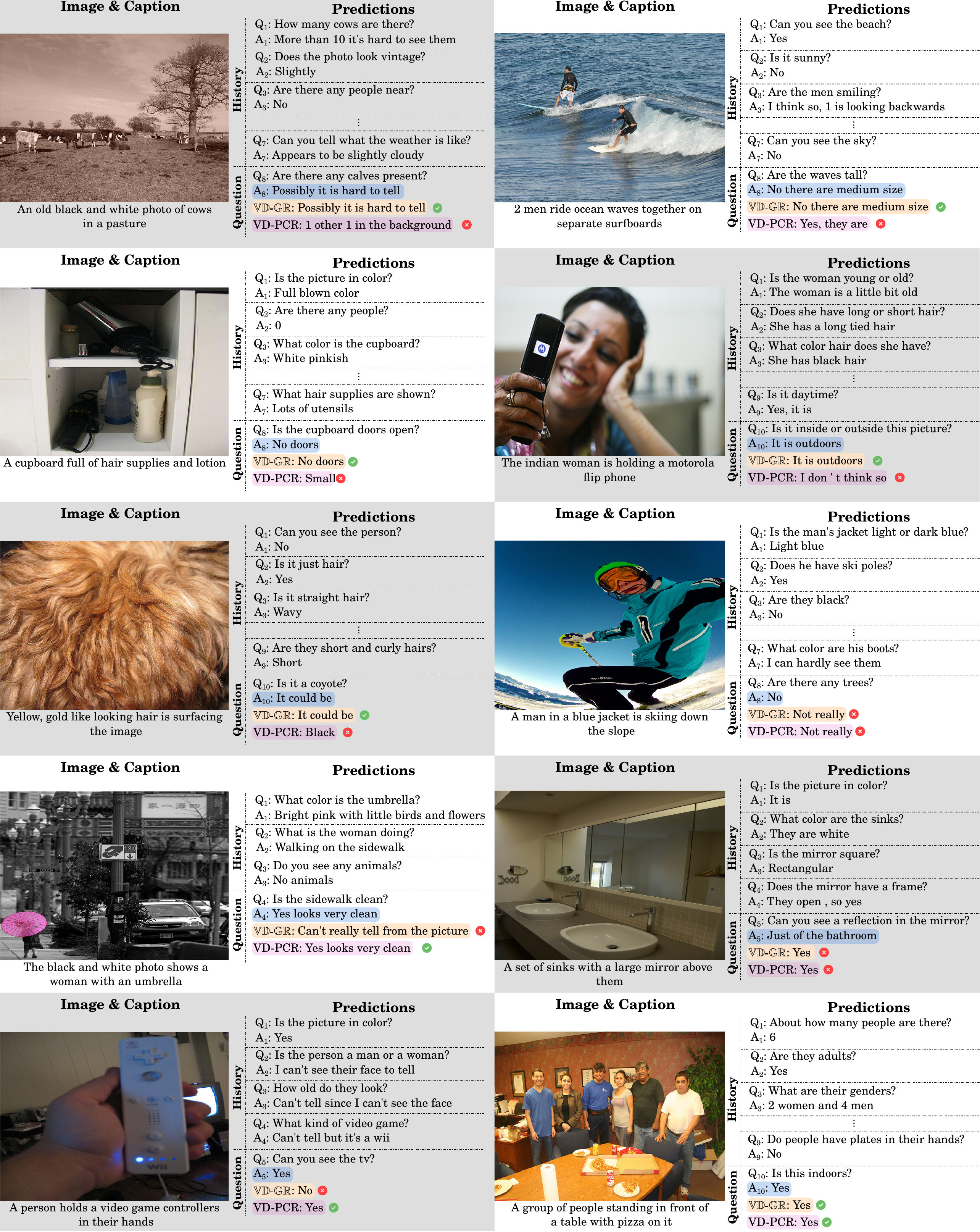}
        }
        \caption{
           Qualitative samples from the \textit{val} split of the VisDial v1.0 dataset.
           We compared the predictions of \vdgr\, (orange) with the ground truth answers (blue) and the predictions of VD-PCR (pink).
           The answers of both models correspond to the top-$1$ predictions.
        }
        \label{fig:qualitative_2}
    \end{minipage}
\end{figure*}

We present additional qualitative examples from the \textit{val} split of VisDial v1.0 in \autoref{fig:qualitative_1} and \autoref{fig:qualitative_2}.
As in the main text, we compared the top-$1$ predictions of \vdgr\, with the ground-truth and the predictions of VD-PCR since it achieved the second best performance on this split.

%%%%%%%%% REFERENCES
{\small
\bibliographystyle{ieee_fullname}
\bibliography{egbib}

\begin{thebibliography}{10}\itemsep=-1pt

\bibitem{Agarwal2020}
Shubham Agarwal, Trung Bui, Joon-Young Lee, Ioannis Konstas, and Verena Rieser.
\newblock {History for Visual Dialog: Do we really need it?}
\newblock In {\em ACL}, 2020.

\bibitem{b2t2}
Chris Alberti, Jeffrey Ling, Michael Collins, and David Reitter.
\newblock {Fusion of Detected Objects in Text for Visual Question Answering}.
\newblock In {\em EMNLP}, 2019.

\bibitem{VQA}
Stanislaw Antol, Aishwarya Agrawal, Jiasen Lu, Margaret Mitchell, Dhruv Batra, C.~Lawrence Zitnick, and Devi Parikh.
\newblock {VQA}: {V}isual {Q}uestion {A}nswering.
\newblock In {\em ICCV}, 2015.

\bibitem{listnet}
Zhe Cao, Tao Qin, Tie-Yan Liu, Ming-Feng Tsai, and Hang Li.
\newblock {Learning to Rank: From Pairwise Approach to Listwise Approach}.
\newblock In {\em ICML}, 2007.

\bibitem{Chen2022}
Cheng Chen, Yudong Zhu, Zhenshan Tan, Qingrong Cheng, Xin Jiang, Qun Liu, and Xiaodong Gu.
\newblock {UTC: A Unified Transformer with Inter-Task Contrastive Learning for Visual Dialog}.
\newblock In {\em CVPR}, 2022.

\bibitem{gog}
Feilong Chen, Xiuyi Chen, Fandong Meng, Peng Li, and Jie Zhou.
\newblock {G}o{G}: Relation-aware graph-over-graph network for visual dialog.
\newblock In {\em Findings of ACL}, 2021.

\bibitem{Chen2020}
Yu Chen, Lingfei Wu, and Mohammed~J. Zaki.
\newblock {Iterative deep graph learning for graph neural networks: Better and robust node embeddings}.
\newblock In {\em NeurIPS}, 2020.

\bibitem{chen2020uniter}
Yen-Chun Chen, Linjie Li, Licheng Yu, Ahmed~El Kholy, Faisal Ahmed, Zhe Gan, Yu Cheng, and Jingjing Liu.
\newblock {Uniter: Universal image-text representation learning}.
\newblock In {\em ECCV}, 2020.

\bibitem{visdial}
Abhishek Das, Satwik Kottur, Khushi Gupta, Avi Singh, Deshraj Yadav, Jos\'e~M.F. Moura, Devi Parikh, and Dhruv Batra.
\newblock {V}isual {D}ialog.
\newblock In {\em CVPR}, 2017.

\bibitem{bert}
Jacob Devlin, Ming-Wei Chang, Kenton Lee, and Kristina Toutanova.
\newblock {BERT}: Pre-training of deep bidirectional transformers for language understanding.
\newblock In {\em NAACL}, 2019.

\bibitem{Elinas2020}
Pantelis Elinas, Edwin~V. Bonilla, and Louis~C. Tiao.
\newblock {Variational inference for graph convolutional networks in the absence of graph data and adversarial settings}.
\newblock In {\em NeurIPS}, 2020.

\bibitem{arxiv.1710.06513}
Haoshu Fang, Yuanlu Xu, Wenguan Wang, Xiaobai Liu, and Song-Chun Zhu.
\newblock {Learning Pose Grammar to Encode Human Body Configuration for 3D Pose Estimation}.
\newblock In {\em AAAI}, 2017.

\bibitem{franceschi2019learning}
Luca Franceschi, Mathias Niepert, Massimiliano Pontil, and Xiao He.
\newblock Learning discrete structures for graph neural networks.
\newblock In {\em ICML}, 2019.

\bibitem{ijcai2019p693}
Dan Guo, Hui Wang, and Meng Wang.
\newblock Dual visual attention network for visual dialog.
\newblock In {\em IJCAI}, 2019.

\bibitem{Guo2020}
Dan Guo, Hui Wang, Hanwang Zhang, Zheng~Jun Zha, and Meng Wang.
\newblock {Iterative Context-Aware Graph Inference for Visual Dialog}.
\newblock In {\em CVPR}, 2020.

\bibitem{resnet}
Kaiming He, Xiangyu Zhang, Shaoqing Ren, and Jian Sun.
\newblock Deep residual learning for image recognition.
\newblock In {\em CVPR}, 2016.

\bibitem{hochreiter1997long}
Sepp Hochreiter and J{\"u}rgen Schmidhuber.
\newblock Long short-term memory.
\newblock {\em Neural computation}, 1997.

\bibitem{rnn}
J~J Hopfield.
\newblock {Neural networks and physical systems with emergent collective computational abilities.}
\newblock {\em Proceedings of the National Academy of Sciences}, 1982.

\bibitem{jiang2020visual}
Tianling Jiang, Yi Ji, Chunping Liu, and Hailin Shao.
\newblock Visual-textual alignment for graph inference in visual dialog.
\newblock In {\em COLING}, 2020.

\bibitem{kbgn}
Xiaoze Jiang, Siyi Du, Zengchang Qin, Yajing Sun, and Jing Yu.
\newblock {KBGN: Knowledge-Bridge Graph Network for Adaptive Vision-Text Reasoning in Visual Dialogue}.
\newblock In {\em ACM MM}, 2020.

\bibitem{jiang2019dualvd}
Xiaoze Jiang, Jing Yu, Zengchang Qin, Yingying Zhuang, Xingxing Zhang, Yue Hu, and Qi Wu.
\newblock {DualVD: An Adaptive Dual Encoding Model for Deep Visual Understanding in Visual Dialogue}.
\newblock In {\em AAAI}, 2020.

\bibitem{student}
Gi-Cheon Kang, Sungdong Kim, Jin-Hwa Kim, Donghyun Kwak, and Byoung-Tak Zhang.
\newblock {The Dialog Must Go On: Improving Visual Dialog via Generative Self-Training}.
\newblock In {\em CVPR}, 2023.

\bibitem{kang2019dual}
Gi-Cheon Kang, Jaeseo Lim, and Byoung-Tak Zhang.
\newblock Dual attention networks for visual reference resolution in visual dialog.
\newblock In {\em EMNLP}, 2019.

\bibitem{kang2020reasoning}
Gi-Cheon Kang, Junseok Park, Hwaran Lee, Byoung-Tak Zhang, and Jin-Hwa Kim.
\newblock Reasoning visual dialog with sparse graph learning and knowledge transfer.
\newblock In {\em Findings of EMNLP}, 2021.

\bibitem{adam}
Diederik~P. Kingma and Jimmy Ba.
\newblock Adam: {A} method for stochastic optimization.
\newblock In {\em ICLR}, 2015.

\bibitem{coref-nmn}
Satwik Kottur, Jos{\'e}~MF Moura, Devi Parikh, Dhruv Batra, and Marcus Rohrbach.
\newblock Visual coreference resolution in visual dialog using neural module networks.
\newblock In {\em ECCV}, 2018.

\bibitem{Kottur2019}
Satwik Kottur, Jos{\'{e}}~M.F. Moura, Devi Parikh, Dhruv Batra, and Marcus Rohrbach.
\newblock {Clevr-dialog: A diagnostic dataset for multi-round reasoning in visual dialog}.
\newblock In {\em NAACL}, 2019.

\bibitem{krishna2017visual}
Ranjay Krishna, Yuke Zhu, Oliver Groth, Justin Johnson, Kenji Hata, Joshua Kravitz, Stephanie Chen, Yannis Kalantidis, Li-Jia Li, David~A Shamma, et~al.
\newblock Visual genome: Connecting language and vision using crowdsourced dense image annotations.
\newblock {\em IJCV}, 2017.

\bibitem{Li_2019_ICCV}
Guohao Li, Matthias Muller, Ali Thabet, and Bernard Ghanem.
\newblock {DeepGCNs: Can GCNs Go As Deep As CNNs?}
\newblock In {\em ICCV}, 2019.

\bibitem{li2019relation}
Linjie Li, Zhe Gan, Yu Cheng, and Jingjing Liu.
\newblock {Relation-aware Graph Attention Network for Visual Question Answering}.
\newblock In {\em ICCV}, 2019.

\bibitem{visualbert}
Liunian~Harold Li, Mark Yatskar, Da Yin, Cho-Jui Hsieh, and Kai-Wei Chang.
\newblock {VisualBERT: A Simple and Performant Baseline for Vision and Language}.
\newblock In {\em arXiv:1908.03557}, 2019.

\bibitem{li-moens-2021-modeling}
Mingxiao Li and Marie-Francine Moens.
\newblock Modeling coreference relations in visual dialog.
\newblock In {\em EACL}, 2021.

\bibitem{arxiv.1511.05493}
Yujia Li, Daniel Tarlow, Marc Brockschmidt, and Richard Zemel.
\newblock {Gated Graph Sequence Neural Networks}.
\newblock In {\em ICLR}, 2015.

\bibitem{arxiv.1506.02108}
Guosheng Lin, Chunhua Shen, Ian Reid, and Anton van~den Hengel.
\newblock {Deeply Learning the Messages in Message Passing Inference}.
\newblock In {\em NeurIPS}, 2015.

\bibitem{lu2019vilbert}
Jiasen Lu, Dhruv Batra, Devi Parikh, and Stefan Lee.
\newblock {ViLBERT: Pretraining Task-Agnostic Visiolinguistic Representations for Vision-and-Language Tasks}.
\newblock In {\em NeurIPS}, 2019.

\bibitem{lu2017best}
Jiasen Lu, Anitha Kannan, , Jianwei Yang, Devi Parikh, and Dhruv Batra.
\newblock {Best of Both Worlds: Transferring Knowledge from Discriminative Learning to a Generative Visual Dialog Model}.
\newblock In {\em NeurIPS}, 2017.

\bibitem{murahari2020large}
Vishvak Murahari, Dhruv Batra, Devi Parikh, and Abhishek Das.
\newblock {Large-scale pretraining for visual dialog: A simple state-of-the-art baseline}.
\newblock In {\em ECCV}, 2020.

\bibitem{Nguyen2020}
Van~Quang Nguyen, Masanori Suganuma, and Takayuki Okatani.
\newblock {Efficient Attention Mechanism for Visual Dialog that Can Handle All the Interactions Between Multiple Inputs}.
\newblock In {\em ECCV}, 2020.

\bibitem{niu2019recursive}
Yulei Niu, Hanwang Zhang, Manli Zhang, Jianhong Zhang, Zhiwu Lu, and Ji-Rong Wen.
\newblock {Recursive Visual Attention in Visual Dialog}.
\newblock In {\em CVPR}, 2019.

\bibitem{pytorch}
Adam Paszke, Sam Gross, Francisco Massa, Adam Lerer, James Bradbury, Gregory Chanan, Trevor Killeen, Zeming Lin, Natalia Gimelshein, Luca Antiga, Alban Desmaison, Andreas Kopf, Edward Yang, Zachary DeVito, Martin Raison, Alykhan Tejani, Sasank Chilamkurthy, Benoit Steiner, Lu Fang, Junjie Bai, and Soumith Chintala.
\newblock {PyTorch: An Imperative Style, High-Performance Deep Learning Library}.
\newblock In {\em NeurIPS}, 2019.

\bibitem{qi2020two}
Jiaxin Qi, Yulei Niu, Jianqiang Huang, and Hanwang Zhang.
\newblock Two causal principles for improving visual dialog.
\newblock In {\em CVPR}, 2020.

\bibitem{qi2020stanza}
Peng Qi, Yuhao Zhang, Yuhui Zhang, Jason Bolton, and Christopher~D. Manning.
\newblock Stanza: A {Python} natural language processing toolkit for many human languages.
\newblock In {\em ACL}, 2020.

\bibitem{NIPS2015_14bfa6bb}
Shaoqing Ren, Kaiming He, Ross Girshick, and Jian Sun.
\newblock {Faster R-CNN: Towards Real-Time Object Detection with Region Proposal Networks}.
\newblock In {\em NeurIPS}, 2015.

\bibitem{Schwartz2019}
Idan Schwartz, Seunghak Yu, Tamir Hazan, and Alexander~G. Schwing.
\newblock {Factor Graph Attention}.
\newblock In {\em CVPR}, 2019.

\bibitem{NIPS2017_654ad60e}
Paul~Hongsuck Seo, Andreas Lehrmann, Bohyung Han, and Leonid Sigal.
\newblock {Visual Reference Resolution using Attention Memory for Visual Dialog}.
\newblock In {\em NeurIPS}, 2017.

\bibitem{hierarchical}
Iulian Serban, Alessandro Sordoni, Ryan Lowe, Laurent Charlin, Joelle Pineau, Aaron Courville, and Yoshua Bengio.
\newblock {A Hierarchical Latent Variable Encoder-Decoder Model for Generating Dialogues}.
\newblock In {\em AAAI}, 2017.

\bibitem{VL_BERT}
Weijie Su, Xizhou Zhu, Yue Cao, Bin Li, Lewei Lu, Furu Wei, and Jifeng Dai.
\newblock {VL-BERT: Pre-training of Generic Visual-Linguistic Representations}.
\newblock In {\em ICLR}, 2020.

\bibitem{sukhbaatar2016learning}
Sainbayar Sukhbaatar, Rob Fergus, et~al.
\newblock Learning multiagent communication with backpropagation.
\newblock In {\em NeurIPS}, 2016.

\bibitem{tan2019lxmert}
Hao Tan and Mohit Bansal.
\newblock {LXMERT: Learning Cross-Modality Encoder Representations from Transformers}.
\newblock In {\em EMNLP}, 2019.

\bibitem{transformer}
Ashish Vaswani, Noam Shazeer, Niki Parmar, Jakob Uszkoreit, Llion Jones, Aidan~N Gomez, {\L}ukasz Kaiser, and Illia Polosukhin.
\newblock {Attention is All you Need}.
\newblock In {\em NeurIPS}, 2017.

\bibitem{gat}
Petar Veličković, Guillem Cucurull, Arantxa Casanova, Adriana Romero, Pietro Liò, and Yoshua Bengio.
\newblock {Graph Attention Networks}.
\newblock In {\em ICLR}, 2018.

\bibitem{Wang2020}
Yue Wang, Shafiq Joty, Michael~R. Lyu, Irwin King, Caiming Xiong, and Steven~C.H. Hoi.
\newblock {VD-BERT: A unified vision and dialog transformer with BERT}.
\newblock In {\em EMNLP}, 2020.

\bibitem{wang2022unified}
Zihao Wang, Junli Wang, and Changjun Jiang.
\newblock Unified multimodal model with unlikelihood training for visual dialog.
\newblock In {\em ACM MM}, 2022.

\bibitem{mem_nets}
Jason Weston, Sumit Chopra, and Antoine Bordes.
\newblock {Memory Networks}.
\newblock In {\em NeurIPS}, 2014.

\bibitem{video_qa}
Jun Xu, Tao Mei, Ting Yao, and Yong Rui.
\newblock {MSR-VTT: A Large Video Description Dataset for Bridging Video and Language}.
\newblock In {\em CVPR}, 2016.

\bibitem{xu2021vitae}
Yufei Xu, Qiming Zhang, Jing Zhang, and Dacheng Tao.
\newblock Vitae: Vision transformer advanced by exploring intrinsic inductive bias.
\newblock In {\em NeurIPS}, 2021.

\bibitem{Yang2021}
Junhan Yang, Zheng Liu, Shitao Xiao, Chaozhuo Li, Defu Lian, Sanjay Agrawal, Amit Singh, Guangzhong Sun, and Xing Xie.
\newblock {GraphFormers: GNN-nested Transformers for Representation Learning on Textual Graph}.
\newblock In {\em NeurIPS}, 2021.

\bibitem{Yao2018}
Ting Yao, Yingwei Pan, Yehao Li, and Tao Mei.
\newblock {Exploring visual relationship for image captioning}.
\newblock In {\em ECCV}, 2018.

\bibitem{ying2021do}
Chengxuan Ying, Tianle Cai, Shengjie Luo, Shuxin Zheng, Guolin Ke, Di He, Yanming Shen, and Tie-Yan Liu.
\newblock Do transformers really perform badly for graph representation?
\newblock In {\em NeurIPS}, 2021.

\bibitem{young-etal-2014-image}
Peter Young, Alice Lai, Micah Hodosh, and Julia Hockenmaier.
\newblock From image descriptions to visual denotations: New similarity metrics for semantic inference over event descriptions.
\newblock {\em TACL}, 2014.

\bibitem{vd_pcr}
Xintong Yu, Hongming Zhang, Ruixin Hong, Yangqiu Song, and Changshui Zhang.
\newblock {VD-PCR: Improving visual dialog with pronoun coreference resolution}.
\newblock {\em Pattern Recognition}, 2022.

\bibitem{vispro}
Xintong Yu, Hongming Zhang, Yangqiu Song, Yan Song, and Changshui Zhang.
\newblock {What You See is What You Get: Visual Pronoun Coreference Resolution in Dialogues}.
\newblock In {\em EMNLP-IJCNLP}, 2019.

\bibitem{yu2017multi}
Zhou Yu, Jun Yu, Jianping Fan, and Dacheng Tao.
\newblock Multi-modal factorized bilinear pooling with co-attention learning for visual question answering.
\newblock In {\em ICCV}, 2017.

\bibitem{zellers2019vcr}
Rowan Zellers, Yonatan Bisk, Ali Farhadi, and Yejin Choi.
\newblock {From Recognition to Cognition: Visual Commonsense Reasoning}.
\newblock In {\em CVPR}, June 2019.

\bibitem{zheng2019reasoning}
Zilong Zheng, Wenguan Wang, Siyuan Qi, and Song-Chun Zhu.
\newblock {Reasoning Visual Dialogs with Structural and Partial Observations}.
\newblock In {\em CVPR}, 2019.

\bibitem{Zhou_Palangi_Zhang_Hu_Corso_Gao_2020}
Luowei Zhou, Hamid Palangi, Lei Zhang, Houdong Hu, Jason Corso, and Jianfeng Gao.
\newblock {Unified Vision-Language Pre-Training for Image Captioning and VQA}.
\newblock In {\em AAAI}, 2020.

\end{thebibliography}
}

\end{document}